\newif\ifARXIV
\newif\ifHAL
\HALtrue

\ifHAL
\documentclass[11pt,a4paper]{article}
\usepackage{natbib}
\usepackage{a4wide}
\else
\documentclass[preprint]{elsarticle}
\fi

\usepackage{moreverb,url}

\usepackage[colorlinks,bookmarksopen,bookmarksnumbered,allcolors=blue]{hyperref}

\ifHAL 
\usepackage[hyperpageref]{backref} 
\renewcommand*{\backref}[1]{}  
\renewcommand*{\backrefalt}[4]{
  \ifcase #1 
  No cited.
  \or
  Cited on page #2.
  \else
  Cited on pages #2.
  \fi}

\usepackage{doi} %maybe put it back in the hal version...
\fi
\usepackage[ruled,vlined]{algorithm2e}
\SetAlFnt{\scriptsize} % or footnotesize

\usepackage{amsmath,amssymb,amsfonts}
\usepackage[utf8]{inputenc}
\usepackage{graphicx}
\usepackage[scriptsize]{subfigure}
\usepackage{xspace}
\usepackage{genom}
\usepackage{listings}
\usepackage{lstlang3}
\usepackage{xcolor}
\usepackage{makecell}
\usepackage{placeins}
\usepackage[binary-units]{siunitx}

\newcommand{\fiacre}{{\sc Fiacre}}
\newcommand{\hfiacre}{{\sc H-Fiacre}}
\newcommand{\tina}{{\sc Tina}}
\newcommand{\hippo}{{\sc Hippo}}
\newcommand{\ps}{{\sc ProSkill}}
\newcommand{\proskill}{{\sc ProSkill}}

\newcommand{\bttf}{{\sc BT2Fiacre}}

\newcommand{\task}[1]{\textsl{#1}}
\newcommand{\event}[1]{\textsl{#1}}
\newcommand{\port}[1]{\textsf{#1}}

\newcommand{\bt}[1]{\textbf{#1}}
\newcommand{\btn}[1]{\textsl{#1}}
\newcommand{\process}[1]{\textbf{#1}}
\newcommand{\sv}[1]{\textsf{#1}}
\newcommand{\svv}[1]{\textit{#1}}
\newcommand{\state}[1]{\textit{#1}}

\newcommand{\code}[1]{{\small\ttfamily{#1}}}
\newcommand{\tool}[1]{\textsf{#1}}

\newcommand{\running}{\code{running}}
\newcommand{\success}{\code{success}}
\newcommand{\failure}{\code{failure}}

\usepackage{inconsolata}
\definecolor{eclipseBlue}{RGB}{42,0.0,255}
\definecolor{eclipseGreen}{RGB}{63,127,95}
\definecolor{eclipsePurple}{RGB}{127,0,85}

\definecolor{colKeys}{rgb}{1,0,0.75}
\definecolor{colKeys2}{rgb}{0,0,1}
\definecolor{colKeys3}{RGB}{63,127,95}
\definecolor{colIdentifier}{rgb}{0,0,0}
\definecolor{colComments}{rgb}{0,0.5,0}
\definecolor{colString}{rgb}{0.6,0.1,0.1} 

\definecolor{bluekeywords}{rgb}{0.13,0.13,1}
\definecolor{greencomments}{rgb}{0,0.5,0}
\definecolor{redstrings}{rgb}{0.9,0,0}

\definecolor{Maroon}{rgb}{0.5,0,0}
\definecolor{darkgreen}{rgb}{0,0.5,0}

\lstdefinelanguage{XML}
{
  basicstyle=\scriptsize\ttfamily,%\ttfamily\footnotesize,
  morestring=[b]",
  moredelim=[s][\color{bluekeywords}]{<}{\ },
  moredelim=[s][\color{bluekeywords}]{</}{>},
  moredelim=[l][\color{bluekeywords}]{/>},
  moredelim=[l][\color{bluekeywords}]{>},
  morecomment=[s]{<?}{?>},
  morecomment=[s]{<!--}{-->},
  commentstyle=\color{colComments},
  stringstyle=\color{colString},
  identifierstyle=\color{colIdentifier}
}

\begin{document}

\renewcommand{\cite}[1]{\citep{#1}}

\ifHAL
\title{A formal implementation of Behavior Trees to act in robotics}%\\
%\textcolor{red}{DRAFT, do no distribute!}}
\else
\title{A formal implementation of Behavior Trees to act in robotics}
\fi

\ifHAL
\author{Félix Ingrand\\
  felix@laas.fr\\
  LAAS-CNRS, Universit\'e de Toulouse\\
  Toulouse, France}
\date{}

\maketitle

\else
\author[1]{Félix Ingrand\corref{cor1}} \ead{felix@laas.fr}
\cortext[cor1]{Corresponding author}
\address[1]{LAAS-CNRS, Universit\'e de Toulouse, Toulouse, France}
\fi

\begin{abstract}
  Behavior Trees (BT) are becoming quite popular as an \emph{Acting} component of autonomous robotic systems. We propose to define a formal
  semantics to BT by translating them to a formal language which enables us to perform verification of programs written with BT, as well as
  runtime verification while these BT execute. This allows us to formally verify BT correctness without requiring BT programmers to master
  formal languages and without compromising BT most valuable features: modularity, flexibility and reusability. We present the formal
  framework we use: \fiacre{}, its language and the produced TTS model; \tina{}, its model checking tools and \hippo{}, its runtime
  verification engine. We then show how the translation from BT to \fiacre{} is automatically done, the type of formal LTL and CTL
  properties we can check offline and how to execute the formal model online in place of a regular BT engine. We illustrate our approach on
  two robotics applications, and show how BT can be extended with state variables, \btn{eval} nodes, node evaluation results and benefit of
  other features available in the \fiacre{} formal framework (e.g., time).
\end{abstract}

\ifHAL
\else

\maketitle 

\fi

\ifHAL
%\clearpage 
%\begin{small}                   %To fit on one page
%\tableofcontents 
%\end{small}
%\clearpage 
\else
%\begin{small}                   %To fit on one page
%\tableofcontents 
%\end{small}
\fi
\section{Introduction and motivation}
\label{sec:intro}
\label{sec:btgeneral}

Behavior Trees (BT) were initially developed and deployed to program Non-Player Characters (NPCs) in video games.  BT are a powerful and
modular framework for designing and implementing decision-making and control systems. They provide a structured way to define complex
behaviors by combining smaller, reusable behavior modules into a hierarchical tree structure. BT have gained popularity in robotics due to
their flexibility, readability, and ability to handle dynamic environments (e.g., Nav2, a popular navigation stack used in ROS2, is now
deployed using BT~\cite{ros2nav2:2024aa}, similarly, BT are a keystone of projects such as~\cite{Street:2024aa} where they aim at verifying the
whole development toolchain).

Why yet another BT implementation? To provide a formal representation of BT which can then be used to formally verify interesting properties
of the BT (can this node succeed, or fail, is this node reachable or not, can we prove that if we reach this state, eventually we will reach
this one within some time interval, etc), and also to execute the formal model which implements the BT in lieu of a regular BT engine.

A behavior tree consists of nodes arranged hierarchically. Parent nodes tick their children to pass them the execution
``token''.  Children return either \success{}, \running{} or \failure{} to their parent when the current execution tick is done, and the
parent pursues the execution following a specified behavior.
\begin{description}
\item[Root node] The entry point of the tree that initiates the behavior evaluation and generates the successive ticks.
\item[\btn{Leaf} nodes] The terminal nodes that perform specific functions when ticked:
  \begin{description}
  \item[\btn{Condition} nodes] Evaluate conditions, such as checking sensor data, battery level, or environmental states.  They return
    \success{} or \failure{}.
  \item[\btn{Action} nodes] Trigger actions in the environment, such as moving a joint, picking up an object, or sending a command. Action,
    on top of \success{} or \failure{}, 
    may also return \running{} (e.g., when it takes \emph{some} time to complete the action such as a robot motion).
  \end{description}
\item[\btn{Control} nodes] These modes specify the execution logic (behavior) of their children nodes (i.e., how to execute them):
  \begin{description}
  \item[\btn{Sequence}] Executes its children from left to right. If any child fails, the \btn{Sequence} fails, halting further
    execution. If the last one succeeds, the \btn{Sequence} succeeds. If a child returns \running{}, so does the \btn{Sequence}, which will,
    when ticked again,
    either call the last running child or restart the \btn{Sequence}, depending on its type: \btn{Sequence} or \btn{ReactiveSequence}.
  \item[\btn{Fallback}] (a mirror of \btn{Sequence}) Executes its children from left to right. If any child succeeds, the \btn{Fallback}
    succeeds, skipping the rest. If a child returns \failure{}, we tick the next child, unless it was the last one, in which case the
    \btn{Fallback} returns \failure{}. When a child returns \running{}, the \btn{Fallback} returns \running{}, and, when ticked again, will
    tick either the last ticked child or the first one depending on its type: \btn{Fallback} or \btn{ReactiveFallback}.
  \item[\btn{Parallel}] Runs/ticks multiple children simultaneously and succeeds or fails based on some success threshold (e.g., either all or some
    of the children must succeed for \success{}).
  \end{description}
\item[\btn{Decorator} nodes] perform a specific operation on a single child (e.g., the \btn{Invert} node returns \success{} when its child returns
  \failure{}, and vice versa and returns \running{} when the child returns \running{}). Here is a  non-exhaustive list of \btn{Decorator}
  nodes: \btn{Repeat}, \btn{ForceFailure}, \btn{ForceSuccess}, \btn{RetryUntilSuccessful}, etc. 
\end{description}

This is a very quick and shallow description of BT, and we invite the reader to check books (e.g., \cite{Colledanchise:2018ub}) and online
tutorials (e.g., \url{https://www.behaviortree.dev/}, which introduces the popular \code{BehaviorTree.CPP}), to get a complete picture of
the BT ``programming'' ecosystem,

Note that the BT specifications are not ``closed''. If most implementations propose the BT nodes described above, some offer additional
\btn{Control} nodes which implement more ``specific'' control algorithms (e.g., Nav2~\cite{ros2nav2:2024aa} proposes \btn{Recovery},
\btn{PipelineSequence}, \btn{RoundRobin}, ...) or additional \btn{Decorator} nodes (e.g., Nav2 proposes \btn{RateController}, etc).  Overall,
BT are making explicit the \emph{control} of the execution of the nodes. However, for the leaves of the tree (\btn{Action} and
\btn{Condition} nodes), the specification remains minimal and silent on some features (e.g. can one pass arguments to nodes? returning values?
asynchronous calls? time taken by the real execution? etc). Nevertheless, most implementations specify how these features are handled (e.g.,
with black board for variables, or in the C/C++ code called to implement them, multi-threading, etc).

Considering that more and more robotics applications, using BT, may be deployed in critical applications (autonomous drones or vehicles, etc)
or in an environment with human (service robots), we need to be able to prove some safety properties on the BT, and to trust their execution will
remain faithful to the programmer's intentions. For this we believe we need to harness some formal models to the BT language and its execution
engine to enable some formal offline and online validation and verification of BT.

\section{State of the art and proposed approach}

We split the state of the art in two parts, on one side, the approaches and the papers concerned with BT in robotic applications, on the other
side, the ones which study formal models jointly with BT.

\subsection{BT in robotics}
\label{sec:soa}

BT are praised for their modularity, readability, scalability, flexibility, robustness, and supposedly being easy to debug and test. Even
if these are questionable and somewhat subjective, one cannot deny their rising success and interest in robotics for autonomous navigation,
human-robot interaction, manipulation tasks and multi-agent systems.

The book~\cite{Colledanchise:2018ub} covers most, if not all, aspects of BT in robotics. They make an extensive presentation of BT, how they
compare to FSM, how they can be linked to the planning activity, etc. They mention the importance of safety and formalism, although not much
is said on formal proof and verification.

In~\cite{Iovino:2022aa} the authors make a comprehensive and large survey of the topic of BT in AI and robotic applications. The existing
literature is described and categorized based on methods, application areas and contributions, and the paper concludes with a list of open
research challenges: explainable AI, human–robot interaction, safe AI, and the combination of learning and BT.

The work presented in~\cite{Marzinotto:2014tg} shows the equivalence between BT and Controlled Hybrid Dynamical Systems.  Similarly the
authors of~\cite{Ogren:2022aa} study how to deploy Behavior Trees in Robot Control Systems, and they propose an interesting formal analysis
regarding convergence and regions of attraction.

The authors of~\cite{Schulz-Rosengarten:2024aa} address one of the BT ``shortsight'' and propose to add a cleaner communication extension,
but lack formalism and proof on the LF.  The input/output mechanism is inspiring.

As for evaluating BT, the authors of \cite{Gugliermo:2024aa} propose a set of metrics (some static, some gathered from real runs), to
evaluate some BT properties, as to evaluate and analyze them.

\paragraph{Implementation considerations}

Deploying BT in robotic applications requires addressing implementation issues which may not be present in Video Game
programming. In~\cite{Colledanchise:2018vt} the authors present an original approach to handle parallelism and concurrency in BT (CBT) with
execution progress and resources management.  In~\cite{Colledanchise:2021aa} , the same authors point out the issues on memory nodes (to
avoid reevaluating), asynchronous action calls, parameters, halt (blocking or not), etc.

\paragraph{BT and planning}
\label{sec:soabtplanning}

In many robotics architectures, BT is considered as the ``acting'' component of the decisional layer~\cite{Ingrand:2015ue}. This is the case for
example in PlanSys2~\cite{Martin:2021aa} (now 
part of the AIPlan4EU platform~\cite{Micheli:2025aa}) where the planner produces plans as BT which can be deployed for plan execution.  It
is also interesting to study how BT may also be extended to perform some planning.  In \cite{Colledanchise:2019vf} the authors propose a
dynamic modification of BT for planning (planning with back chaining), so they can perform robust acting, without resorting to
replanning. On another yet different type of planning/BT interaction, in~\cite{Kockemann:2023aa} planning is used to produce testing plans
for BT, whose testing participates to increase the  trust we  put in these BT.

\subsection{BT and formal models}
\label{sec:soabtfm}

This paper main subject is about BT and formal V\&V, so we now examine the state of the art in this area.

In~\cite{Klockner:2013aa} the authors propose to interface BT mission plans and a simulation of the world using the description logic
($\mathcal{ALC(D)}$).  So the description logic formal model acts as a safety check between the plan execution, and the simulation.  Even if
this does not provide a formal proof of the mission plans BT, it improves the trust we can have in these plans by formally validating their
execution (in simulation) before deploying them in the real world.

The authors of \cite{Colledanchise:2017ac} propose to synthesize correct by construction BT from an environment specification along the
agent model and an objective expressed in LTL.  From a standpoint, the approach is clearly sound and synthesizes correct BT, but requires
the programmer to write LTL goal specifications to get started which may be seen as a deterrent to non formal ``programmers''.

\vspace{0.5em}

The last four approaches we present here have strong similarities with the one we propose.

In~\cite{Biggar:2020aa} the authors propose to synthesize LTL from  BT and then show that the obtained model satisfies some   LTL
specifications.  The paper goes in depth to explain the translation process, although it is not clear it can be automated, and a priori, the
produced formal model cannot be directly executed in place of a regular BT engine.

The authors of~\cite{Colledanchise:2021ab} focus on the formalisation of the  execution context of BT to be able to perform runtime
verification. They propose \emph{channel systems} to model the ``surroundings'' of the BT and then to check at runtime that some
specifications, written in the SCOPE language, are satisfied while the BT executes. So the approach, which is not limited to BT,  provides a
very strong and formal execution framework sitting between the BT and the robot, behaving like a safety bag. The battery example they present
has some similarity with the one we deploy on our UAV in section~\ref{sec:uav-bt}.

Similarly, the authors of~\cite {Serbinowska:2024ab} focus primarily on runtime verification of BT with contingency monitors (BTM) written with a DSL:
BehaVerify.  These monitors can be used to correct an undesirable behavior when it is detected and can handle LTL specifications. Yet, they
can also check the BT at design time, by checking these BTM with model checking.

In~\cite{Wang:2024aa}, the authors present an approach where they use the BIP formal framework to model BT and propose an implementation of
their tool: \tool{xml2bip}. They then use model checking (not D-Finder as the original BIP implementation did) to check for formal
properties. Although some versions of BIP come with a runtime engine (e.g., the one used in~\cite{Bensalem:2011uf}), they do not yet propose
a ``fornal'' execution of the BT with the BIP engine.

\subsection{Proposed approach}

Our approach has one main goal: to provide a formal semantics for BT, by translating it to a formal model, which can then be used offline to
check formal properties, but also online to implement and enforce this semantics.

We propose to achieve this objective by following these steps.

\begin{itemize}
\item Define a clear complete and unequivocal translation of all BT to a formal model in \fiacre{}~\cite{Berthomieu:2007ab,Berthomieu:2008vo}.
\item The obtained BT formal model can be checked and analyzed to prove logical and temporal properties (LTL and CTL).
\item The \emph{same}  BT formal model can be linked to actual code and executed like other BT framework engines (e.g. \code{Behaviortree.CPP},
  \code{BT.py}) do. This shows that the operational semantic of the BT formal model is the expected one, while guaranteeing the property
  proven offline.
\end{itemize}

Moreover, implementing this approach leads to some interesting side effects and features.  It clarifies the BT semantics when needed, e.g.,
the wait/halt semantics when \running{} nodes must be halted. It also enables time representation extensions and enriches
the BT language with state variables and functions evaluation.

\vspace{1em}

The rest of the paper is organized as follows. After introducing above the BT, the state of the art and our approach, we first present in
Section~\ref{sec:formalfw} the \fiacre{}/\tina{}/\hippo{} formal suite we use.  Section~\ref{sec:btformal} introduces how each BT node is mapped in
\fiacre{} (this is implemented in our \bttf{} tool\footnote{The \bttf{} tool developed for this study can be downloaded from the repository:
  \url{https://redmine.laas.fr/projects/bt2fiacre/pages/index}.}). Then Section ~\ref{sec:deploy} presents how these \fiacre{} BT nodes are
put together to build the complete formal model of the whole BT, and we then show what are the type of formal properties one can prove
offline but also at runtime. Two examples are presented: in Section~\ref{sec:uav-bt}, we introduce a drone controller written in BT for
which we successfully deployed our approach, and; in Section~\ref{sec:nav2}, we show how the Nav2 BT~\cite{ros2nav2:2024aa} can be deployed
with \bttf{}. A discussion in Section~\ref{sec:conclusion} reassesses the pros and cons of the \bttf{} tool and the use of \fiacre{} as an
underlying formal language to provide a formal model and a formal semantics to BT, followed by future work section and the conclusion of the
paper.

\section{A Formal Framework for Offline and Runtime Verification: The \fiacre{} Language, Models, and Tools}
\label{sec:formalfw}

While this paper does not aim to exhaustively present the formal framework we use, some terminology and explanations are essential for
clarity and make the paper self contained. Readers interested in more details may consult the specific papers and websites cited below.
\footnote{Note that a similar \fiacre{} presentation can be found in this paper~\cite{Ingrand:2024aa} (from the same author). We include it
  almost as is in this paper as to make the paper self contained, nevertheless, If the reviewers believe this section should be shortened and
  replaced by a pointer to the other paper, this is perfectly fine for us.}

\subsection{Terminology, Models, Languages, and Tools}

We define the following terms:

\begin{description}

\item[Time Petri Nets] \cite{Berthomieu:1991wv} are an extension of traditional \emph{Petri nets} where each transition has an associated
  time interval (typically $[0,\infty)$) specifying the time range within which an enabled transition can be fired.

\item[TTS] Time Transition Systems extend \emph{Time Petri nets} by adding data-handling capabilities, allowing transitions to invoke data
  processing functions.

\item[\tina{}] (short for "TIme Petri Net Analyzer") is a toolkit for editing, simulating, and analyzing \emph{Petri nets}, \emph{Time Petri
    nets}, and \emph{TTS}. Within this toolkit, \tool{sift} and \tool{selt} enable the construction of reachable state sets and the
  verification of LTL properties.\footnote{\url{https://projects.laas.fr/tina/index.php}}

\item[\fiacre{}] stands for "Intermediate Format for Embedded Distributed Component Architectures" (in French). It is a formally defined
  language designed to represent the behavioral and timing aspects of embedded and distributed systems for purposes of formal verification
  and simulation. \fiacre{} specifications can be compiled into a \emph{TTS} using the \texttt{frac}
  compiler.\footnote{\url{https://projects.laas.fr/fiacre/index.php}}

\item[\hfiacre{}] adds \emph{Event Ports} and \emph{Tasks} linked to C/C++ functions, to make the \fiacre{} models “executable”.
  
\item[\hippo{}] is an engine for executing \emph{TTS} resulting from the \hfiacre{}
  specifications compilation~\citep{Hladik:2021vt}.\footnote{\url{https://projects.laas.fr/hippo/index.php}}

\end{description}

This framework has been applied across various projects and applications,\footnote{\url{https://projects.laas.fr/fiacre/papers.php}}
including the validation and verification of functional components in our robotics experiments~\cite{Dal-Zilio:2023aa}, but also to the
validation and verification of robotic skills programmed in \proskill{}~\cite{Ingrand:2024aa}.

\subsection{\fiacre{} Semantics}
\label{sec:fiacreprocess}

\begin{figure*}[!ht]
\begin{center}
\includegraphics[width=0.75\textwidth]{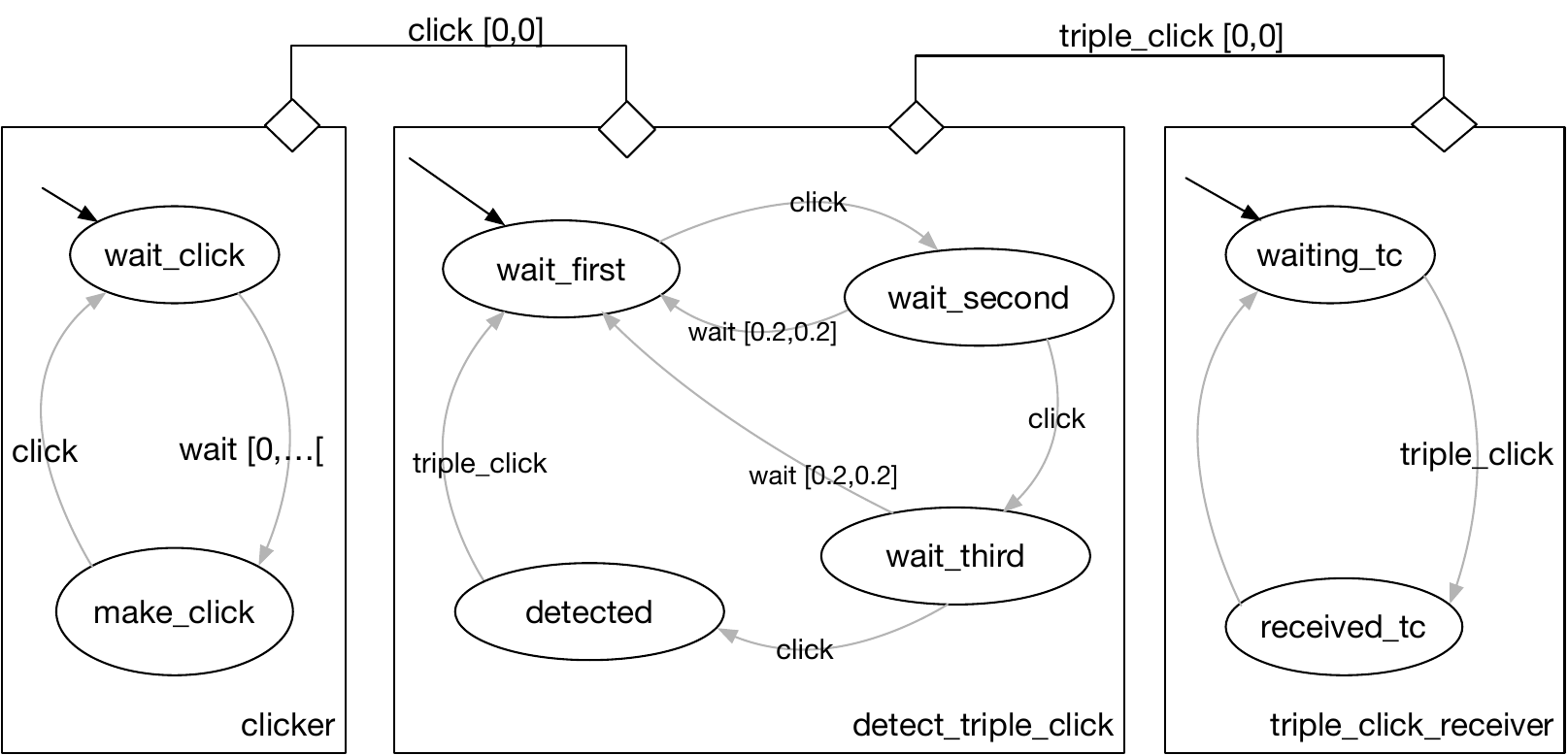}
\caption{The \fiacre{} processes modeling the \fiacre{} specification on Listing~\ref{lst:ftct}\code{p}\pageref{lst:ftct}.}
\label{fig:tcd-tina}
\end{center}
\end{figure*}

Although the formal model and tools are detailed in specific papers and websites (see above), we include a brief example to illustrate the
semantics of the \fiacre{} language. The example, a triple-click detector for a mouse, is shown in
Listing~\ref{lst:ftct}\code{p}\pageref{lst:ftct}\footnote{All floating Listing and Figures numbers are given with the page number. In this case,
  Listing~\ref{lst:ftct}\code{p}\pageref{lst:ftct} is Listing~\ref{lst:ftct}, page~\pageref{lst:ftct}.} and illustrated in
Figure~\ref{fig:tcd-tina}. It defines three \fiacre{} processes, each represented by an automaton. The first process, \process{clicker},
generates a \event{click} at any time, waiting between $0$ and $\infty$, then synchronizes on the \port{click} port with
\process{detect\_triple\_click}. This second process has four states, waiting for synchronization on \port{click} or until the maximum
allowed time between clicks (\qty{0.2}{\sec}) has passed. Note the \code{select} option in \state{wait\_second} and \state{wait\_third}
states, introducing a non-deterministic choice for exploration by the model checker. Upon reaching \state{detected}, a synchronization on
\port{triple\_click} enables the transition of the \process{triple\_click\_receiver} process to \state{received\_tc}.

Following these specifications, a component is defined by placing three process instances in parallel (line~\ref{lst:ftct}.\ref{ll:par})\footnote{Listing lines are referenced with the $<$listing
  number$>$.$<$line number$>$, example:~\ref{lst:ftct}.\ref{ll:par} is Listing~\ref{lst:ftct}, line:~\ref{ll:par}.} and
linking them through two ports (line~\ref{lst:ftct}.\ref{ll:ports}). This example is simple by design, though the \fiacre{} language
supports complex data types, bidirectional ports, local and global variables, conditions, switch/case statements, transition guards, and
function calls (internal to \fiacre{} or external in C/C++ code) for advanced computation. More complex \fiacre{} specifications can be
found in \ref{app:app1} and \ref{app:app2}.

\begin{lstlisting}[caption={\fiacre{} specification for a triple click detector (\fiacre{} offline version).}, numbers=left, xleftmargin=15pt, label={lst:ftct}, language=fiacre]
process clicker [click:sync] is // synthesize clicks and sync them on its port at any time
states wait_click, make_click

from wait_click
   wait [0, ...[; // wait any time from zero to infinity
   to make_click

from make_click
   click; // issue a click sync on the Fiacre port
   to wait_click

process detect_triple_click [click:sync,triple_click:sync] is
states wait_first, wait_second, wait_third, detected

from wait_first
   click;         // first click  (*@ \label{ll:click1} @*) 
   to wait_second

from wait_second
   select        // we wait either
     wait [0.2,0.2]; //  exactly 0.2 second
     to wait_first // then reset the detector
   []
     click;       // or for the second click   (*@ \label{ll:click2} @*) 
     to wait_third // whichever comes first
   end

from wait_third
   select        // again for the third click
     wait [0.2,0.2];
     to wait_first
   []
     click;       // third  (*@ \label{ll:click3} @*) 
     to detected
   end

from detected
   triple_click; // sync on the triple_click port
   to wait_first

process triple_click_receiver[triple_click:sync] is
states waiting_tc, received_tc

from waiting_tc
   triple_click; // just wait for a sync on this port
   to received_tc

from received_tc
   /* do what needs to be done when a TC has been detected */
   to waiting_tc

component comp_tc is //we now specifiy the component

port click:sync in [0,0], triple_click:sync in [0,0] // two ports  (*@ \label{ll:ports} @*) 

par * in // 3 processes composed in parallel  (*@ \label{ll:par} @*) 
   detect_triple_click[click, triple_click] // process 1
|| clicker[click]   // process 2
|| triple_click_receiver[triple_click] //process 3
end

comp_tc // this instantiates the component

// some properties to check
property ddlf is deadlockfree  // deadlock free (TRUE) (*@ \label{ll:ddf} @*) 
assert ddlf
// in the next property comp_tc/3/state designates the state in the 3rd process of
property cannot_receveice_tc is absent comp_tc/3/state received_tc // the comp_tc component (*@ \label{ll:rtc} @*) 
assert cannot_receveice_tc // we cannot detect a triple click (FALSE)
\end{lstlisting}

\subsection{Offline Formal Verification}

The \tool{frac} compiler is used to compile the \fiacre{} specifications shown in Listing~\ref{lst:ftct}\code{p}\pageref{lst:ftct} into an equivalent TTS. With the
\tool{sift} tool from the \tina{} toolbox, one can then construct the system's set of reachable states, and by using \tool{selt}, the
properties outlined in the initial \fiacre{} specifications, as well as any additional properties, can be verified. The \tina{} toolbox
offers a variety of other tools that readers may explore for further analysis.

Listing~\ref{lst:ftct}\code{p}\pageref{lst:ftct} suggests several properties to check for this specification: Is the model deadlock-free
(line~\ref{lst:ftct}.\ref{ll:ddf})? This is confirmed to be TRUE. Can the model successfully detect a triple click
(line~\ref{lst:ftct}.\ref{ll:rtc})? By verifying that reaching the \state{received\_tc} state is possible, the model confirms that it can
indeed detect a triple click. Additional complex properties could be added, such as ensuring there is at most \qty{0.4}{\sec} between the
first and last clicks.

The verification approach used by the \tina{} tools relies on model checking, which can be affected by state explosion~\cite{Clarke:2012uv},
potentially limiting its effectiveness. However, as demonstrated in section~\ref{sec:results}, the results from our example remain both
insightful and non-trivial.

\subsection{\hfiacre{} Runtime Extensions}

Although \fiacre{} was originally designed for offline verification, it has been extended with two primitives that enable runtime
verification~\citep{Hladik:2021vt}. These extensions allow the model to connect with C/C++ functions that send events or execute commands,
forming what we call \hfiacre{}, to distinguish it from the base \fiacre{} language.

The purpose of the \hfiacre{} runtime version is to make the model "executable" in connection with real-world interactions.

\begin{figure*}[!ht]
\begin{center}
\includegraphics[width=0.97\textwidth]{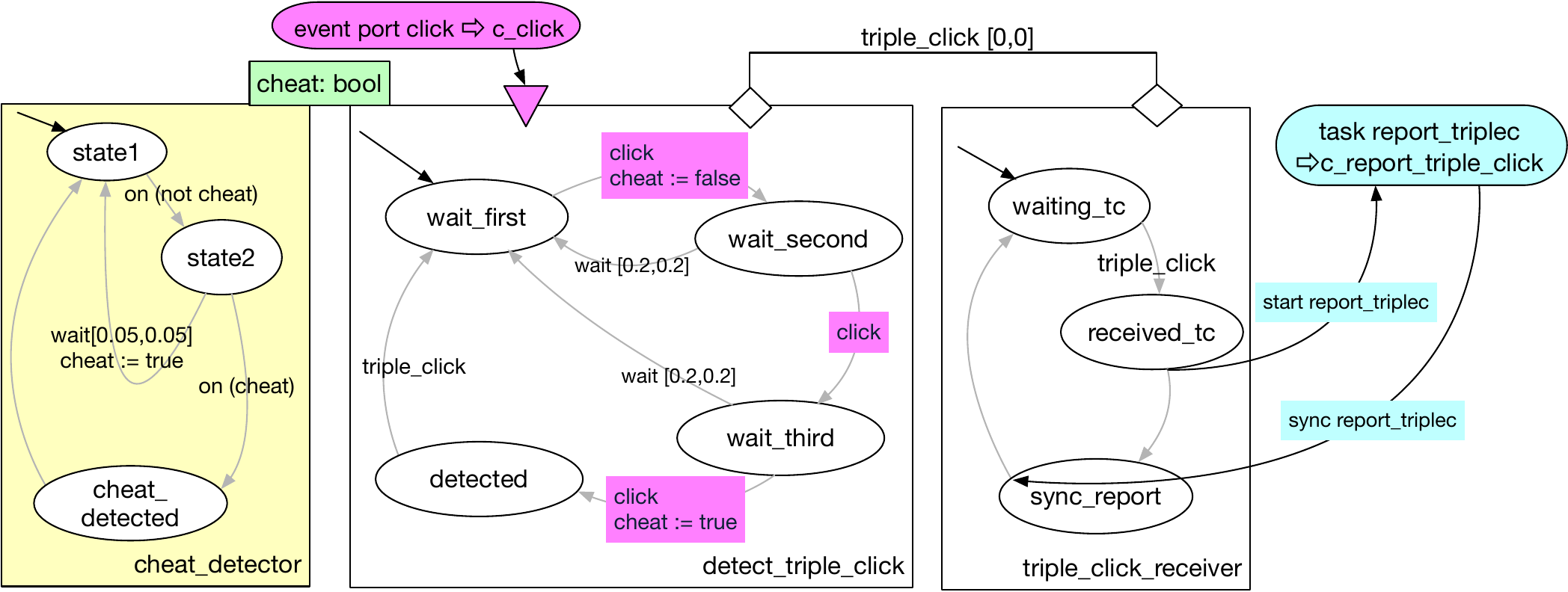}
\caption{Illustration of the \hfiacre{} program on Listing~\ref{lst:ftch}\code{p}\pageref{lst:ftch}.}
\label{fig:tcd-hippo}
\end{center}
\end{figure*}

Listing~\ref{lst:ftch}\code{p}\pageref{lst:ftch} (along with Figure~\ref{fig:tcd-hippo}\code{p}\pageref{fig:tcd-hippo}) presents the executable version of the specification given in Listing~\ref{lst:ftct}\code{p}\pageref{lst:ftct}.

\begin{description}
\item[Event Ports] are defined in the specification's preamble (see line~\ref{lst:ftch}.\ref{ll:event_port}), linking a C function to a
  \fiacre{} port. In this case, the event \event{click} is linked to the \code{c\_click} function in C/C++. When this port is one of the
  possible transitions (lines~\ref{lst:ftct}.\ref{ll:click1},~\ref{lst:ftct}.\ref{ll:click2}, and~\ref{lst:ftct}.\ref{ll:click3}), the C/C++
  function is called, and the port becomes active upon the function's return. These C/C++ functions can accept and return values typed in
  \fiacre{}. 
\item[Tasks] are also defined in the preamble (see line~\ref{lst:ftch}.\ref{ll:task}), associating a task (in this case,
  \task{report\_triplec}) with a C/C++ function (here \code{c\_report\_triple\_click}), which is called asynchronously upon a \code{start}
  (see line~\ref{lst:ftch}.\ref{ll:start}). This enables the corresponding \code{sync} (line~\ref{lst:ftch}.\ref{ll:sync}) once the C/C++
  function completes. Values can be passed to the task at call time and returned when it completes. 
\end{description}

\begin{lstlisting}[caption={\hfiacre{} processes implementing a triple click detector.}, numbers=left, xleftmargin=15pt, label={lst:ftch}, language=fiacre]
event click : sync is c_click // declare the Fiacre event port which transmits click (*@ \label{ll:event_port} @*) 
task report_triplec () : nat is // declare the task and 
	c_report_triple_click  // the C/C++ function called by this task  (*@ \label{ll:task} @*) 

process detect_triple_click [triple_click:sync] is 
// this process is exactly the same than in the regular Fiacre version
// only the click port is now an event port

process triple_click_receiver[triple_click:sync] is
states waiting_tc, received_tc, sync_report
var ignore : nat

from waiting_tc
   triple_click;
   to received_tc

from received_tc // show an example of an external call
   start report_triplec();  (*@ \label{ll:start} @*) 
   to sync_report

from sync_report
   sync report_triplec ignore;  // wait until the call return (*@ \label{ll:sync} @*) 
   to waiting_tc

component comp_tc is
port triple_click:sync

par * in
   detect_triple_click[triple_click]
|| triple_click_receiver[triple_click]
end

comp_tc
\end{lstlisting}

In this example, we replace the \process{clicker} process, which previously synchronized with \port{click} at any moment, with the
\event{click} event port (highlighted in purple). Additionally, we introduce a task (\task{report\_triplec} shown in light blue) to execute
when synchronizing with a \event{triple\_click} in the \process{triple\_click\_receiver} process. The remainder of the model remains
unchanged, transforming it from a model specifying a triple-click detector to an actual program or controller that implements it. In this
way, the specification itself becomes an executable program.

\subsection{Runtime (Online) Verification}
\label{sec:rv}

The \hfiacre{} model, once compiled into TTS format using \tool{frac}, is linked with the \hippo{} engine and the C/C++ functions needed to
run it (e.g., \code{c\_click} and \code{c\_report\_triple\_click}). The \hippo{} engine executes the TTS model in real-time and initiates
the appropriate C/C++ function calls (in separate threads) connected to event ports and tasks. Note that the properties checked offline also
hold in the online version (as the reachable states set of the former include the one form the latter). 

Additionally, the model can be extended with runtime verification properties by incorporating a monitoring process. For instance, if this
controller is applied in a video game where rapid sequences of triple clicks are suspicious (indicating potential use of a cheating device),
a new \process{cheat\_detector} process could be added (shown on the left in Figure~\ref{fig:tcd-hippo}\code{p}\pageref{fig:tcd-hippo} and in Listing~\ref{lst:fcd}\code{p}\pageref{lst:fcd}). This
process could have three states and use a shared Boolean variable, \code{cheat}. This variable would be set to `false` upon transitioning to
the \state{waiting\_second} state in the \process{detect\_triple\_click} process, and switched to `true` upon reaching the \state{detected}
state if the click sequence timing suggests non-human activity.

\begin{lstlisting}[caption={\process{cheat\_detector} process detecting a cheating device by monitoring the \code{cheat} Boolean variable.}, numbers=left, xleftmargin=15pt, label={lst:fcd}, language=fiacre]
process cheat_detector(&cheat:bool) is

states state1, state2, cheat_detected

from state1
  on (not cheat); // guard on (not cheat)
  to state2
  
from state2 // cheat was set to false
  select // either 
     wait [0.05,0.05]; // 50 ms elapsed
     cheat := true; // reset the cheat variable
     to state1 // go back to monitoring
  []
     on (cheat); // cheat became true again before the 50ms above.
     to cheat_detected //caught cheating
  end

from cheat_detected
  // the player is cheating, do what needs to be done.
  to state1
\end{lstlisting}

Within the \process{cheat\_detector} process, in \state{state1}, the system sets a guard on \code{(not cheat)} before transitioning to
\state{state2}, where it then waits for either \qty{50}{\ms} or until \code{cheat} becomes true. If \code{cheat} becomes true before
\qty{50}{\ms} has elapsed (indicating a suspiciously fast triple-click), it transitions to \state{cheat\_detected} and flags this unusual
activity. If \qty{50}{\ms} passes without the \code{cheat} variable being set to true, the system sets \code{cheat} to true and returns to
\state{state1}.

This approach allows us to synthesize a controller that directly runs the specification. This dual capability is a major strength of the
\fiacre{} framework: the same formal model can be verified offline and executed online. In practical terms, this means that the controller’s
real-time behavior aligns with the initial model specifications, validating that the offline-verifiable properties are applied consistently
in the live system. While observing expected behavior is a necessary, though not entirely sufficient, indicator of correctness, it
significantly strengthens the link between specification and execution.

Moreover, if runtime behavior diverges from expectations, you can debug it as you would with any programs. From a formal perspective, the possible
traces of the \hfiacre{} version (also called the \hippo{} version) are contained within those of the \fiacre{} model
(also called the \tina{} version), ensuring consistency between the
runtime model and its  offline counterpart.

\section{The mapping of BT in \fiacre{}}
\label{sec:btformal}

\begin{figure*}[!ht]
\begin{center}
\includegraphics[width=0.97\textwidth]{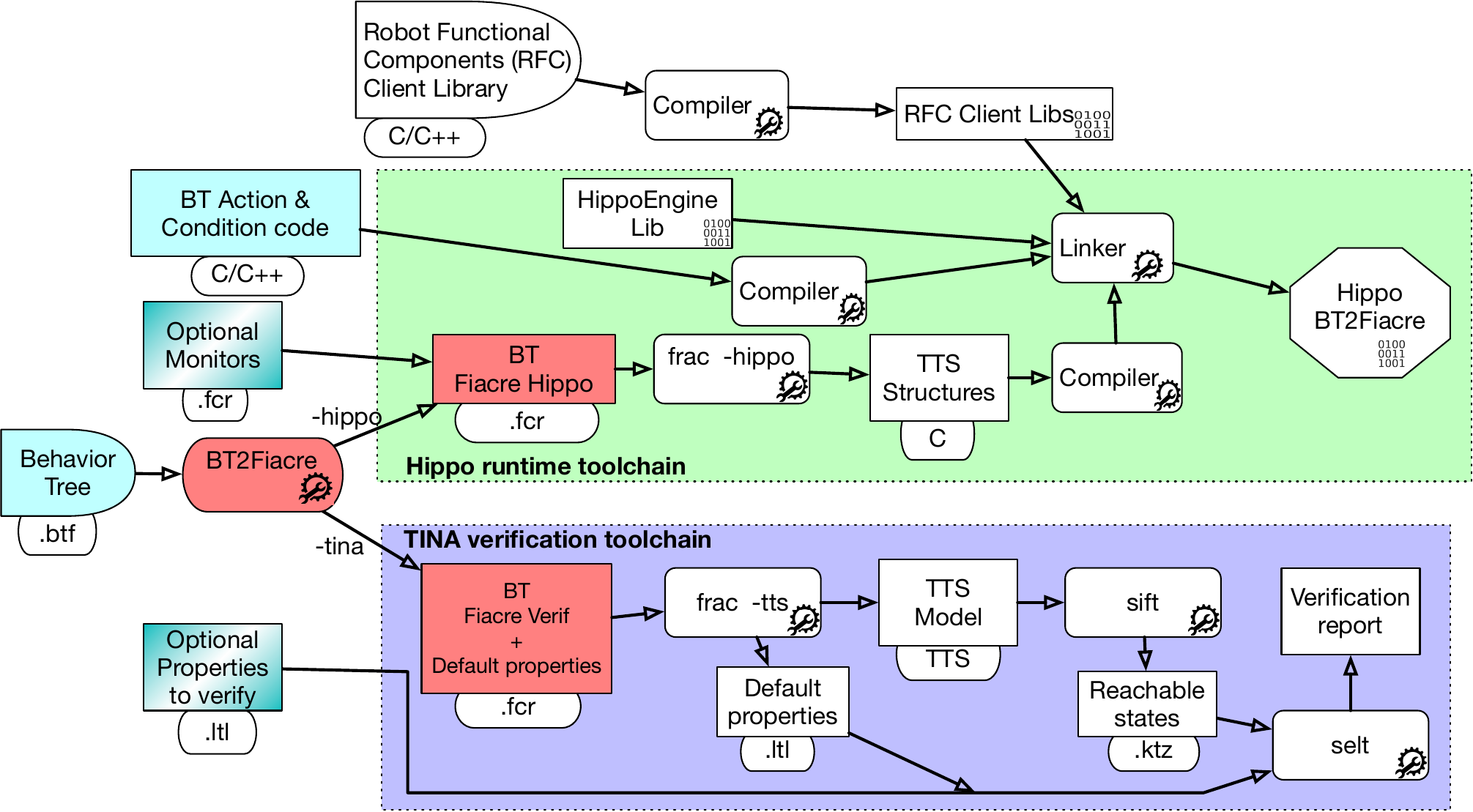}
\caption{The \bttf{} (\hippo{}/\tina{}) workflow. Only the data in the blue boxes need to be provided by the programmer. The \bttf{} tool (in light
  red) synthesizes the two models, the rest is fully automated.  In light green, the workflow for the \hippo{} runtime verification
  version, and in light purple, the workflow for the \tina{} offline verification version.}
\label{fig:workflow}
\end{center} 
\end{figure*}

Before getting into the details of the produced formal models, we present on Figure~\ref{fig:workflow}\code{p}\pageref{fig:workflow} the
overall workflow from BT to the formal executable version (top part in green), and the formal verifiable version and its analyzed properties
report (bottom part in purple).  One should keep in mind, that the BT programmers only provide the various BT in \code{.btf} format (like
the one on Listing~\ref{lst:dronebt}\code{p}\pageref{lst:dronebt}), the C/C++ codes which glue \btn{Action} and \btn{Condition} BT to the
real robot commands and perception primitives (all in blue) and, optionally, LTL properties to verify, and monitors written in \fiacre{} (in
slanted blue). The rest is fully synthesized and automatically compiled.

\begin{lstlisting}[caption={A simple  drone survey  BT. Note that for historical reasons and as we are reusing some of our existing tools, we do not use an XML syntax but rather a Lisp like syntax (called the \code{.btf} format), but this has no consequence on the content and the interpretation of BT (See Appendix~\ref{app:ros2nav2}, Listing~\ref{lst:ros2nav2xml}\code{p}\pageref{lst:ros2nav2xml} and~\ref{lst:ros2nav2btf}\code{p}\pageref{lst:ros2nav2btf} for an example in both formats).}, numbers=left, xleftmargin=15pt, label={lst:dronebt}, language={[btf]Lisp}]
((BehaviorTree :name drone
  (Sequence
   (ParallelAll :wait 1 :halt 0 ; if wait is 1, will wait the running branch if one fails
    (Action :ID start_drone)
    (Action :ID start_camera))
   (ReactiveSequence
    (Fallback
     (Condition :ID battery_ok)  ; check if the battery is OK
     (ForceFailure :ID fail  ; if not, just land and fail
      (Action :ID land)))
    (Fallback
     (Condition :ID localization_ok) ;same for localization
     (ForceFailure :ID fail
      (Action :ID land)))
    (Sequence
     (Action :ID takeoff :args (height 1.0 duration 0)) 
     (Parallel :success 1 :wait 0 :halt 1 ; If the survey or the nav succeed, we are done.
      (Action :ID camera_survey)
      (Sequence
       (Action :ID goto_waypoint :args (x -3 y -3 z 5)) 
       (Action :ID goto_waypoint :args (x -1.5 y 3 z 5))
       (Action :ID goto_waypoint :args (x 0 y -3 z 5))
       (Action :ID goto_waypoint :args (x 1.5 y 3 z 5))
       (Action :ID goto_waypoint :args (x 3 y -3 z 5))
       (Action :ID goto_waypoint :args (x 3 y -3 z 5))))
     (Action :ID goto_waypoint :args (x 0 y 0 z 5))
     (Action :ID land)
     (Action :ID shutdown_drone))))))
\end{lstlisting}

As mentioned above, the semantics of BT is not formally defined. By translating BT to \fiacre{}, we define a formal semantics, hopefully
consistent with the operational semantics people expect from BT.

Another goal we pursue is to model as much BT ``additional information'' as possible in the \fiacre{} model (for example if the variable used in
BT can be modelled in \fiacre{}, the better). 

We shall first present the overall mapping and then we will point out where the semantics had to be clarified and what are the
``additional'' information we gather in the \fiacre{} model.

\subsection{The general mapping}

Each BT node is automatically mapped in a \fiacre{} process (See Section~\ref{sec:fiacreprocess}), whose automata is modeled in accordance to its node type (\btn{control},
\btn{decorator}). Then, all these \fiacre{} processes are instantiated and composed in a component which provides a shared array of \fiacre{}
records. This array \code{BTnode[]} has one element for each BT and is a shared variable between all the \fiacre{} processes. In the following figures \code{BTnode[self]} is
the record for the current BT node. Each \code{BTnode[]} record has two fields of interest: \code{caller} and \code{rstatus}. \code{caller} is
initialized at \code{None} and will be set to the BT node which ticks/calls it. It is set back to \code{None} when its execution (for the
current tick) is finished. \code{rstatus} contains the last returned status (\success{}, \failure{} or \running) of the BT
node.  \code{rstatus} may also be set to \code{halt\_me} (by its caller), when a \running{} BT node must be halted.

\begin{figure*}[!ht]
\begin{center}
\includegraphics[width=0.7\textwidth]{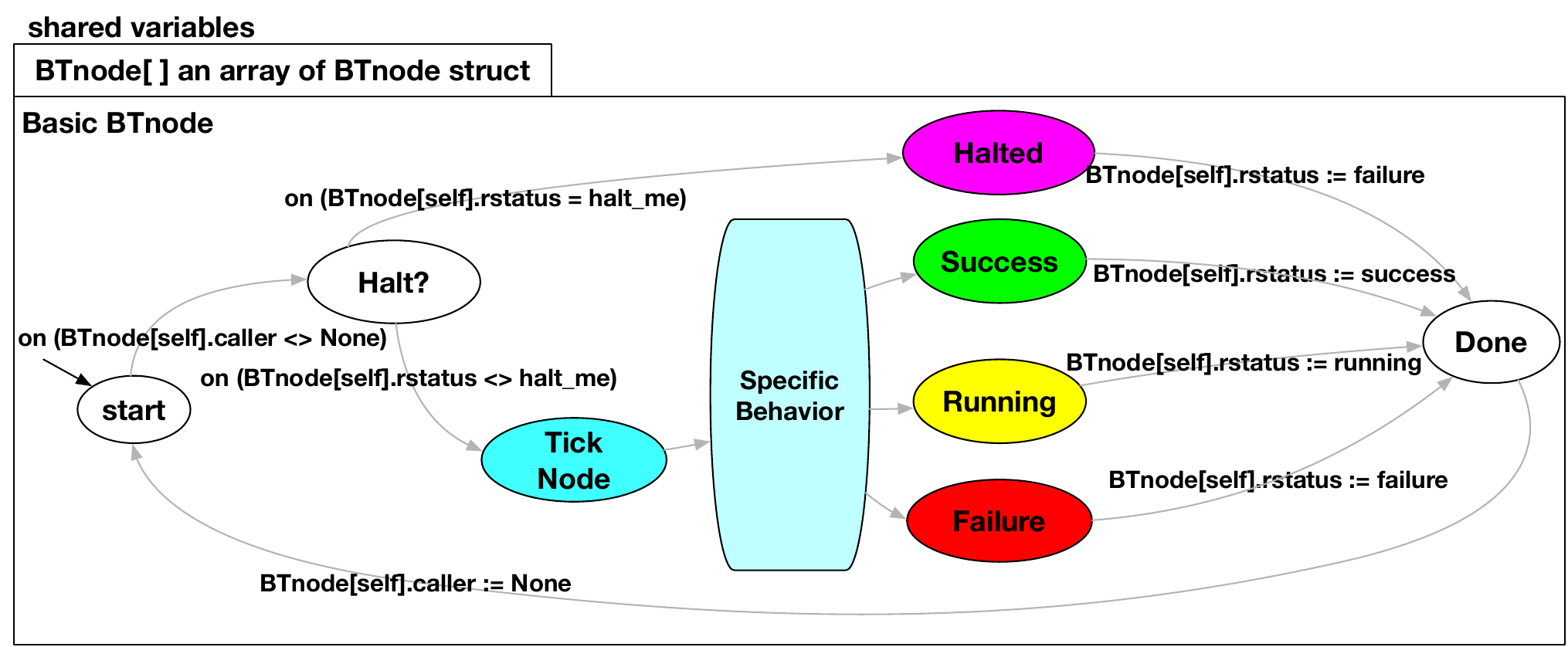}
\caption{Preamble and postamble of BT nodes in \fiacre{}.}
\label{fig:node}
\end{center}
\end{figure*}

As shown on Figure~\ref{fig:node}\code{p}\pageref{fig:node}, a BT node \fiacre{} process %, except the one modelling the root of the BT,
starts with a preamble (on Listing~\ref{lst:factionh}\code{p}\pageref{lst:factionh} or~\ref{lst:fsequenceh}\code{p}\pageref{lst:fsequenceh}
from \state{start\_} to \state{tick\_node}) with a guard on its activation on \code{BTnode[self].caller} (i.e. presumably its parent node
has ticked its by setting its \code{caller} field). Follow a test to check if it has been asked to halt (in case it had previously returned
\running{} and now its parent asks it to halt). If it needs to run, it transitions to the \state{tick\_node} state where the particular of
this node type is handled.

Similarly, the postamble mostly consists of the four automata states (\success{}, \failure{}, \running{} and \code{halted}), all returning the proper
return status \code{rstatus} and then transitioning to a \state{done} state, in which the control is relinquished by setting \code{caller}
to \code{None} (the parent node has a guard on this to check that the node has finished this tick).

In the following \btn{Action} and \btn{Condition} nodes in \fiacre{} are presented in their \tina{} version (i.e., the offline verification) but also \hippo{}
version, (i.e. the runtime version). All other BT nodes \fiacre{} models are strictly the same between the two versions.

\subsection{\btn{Condition} Node}
\label{sec:btncondition}

\begin{figure*}[!ht]
\begin{center}
\includegraphics[width=0.7\textwidth]{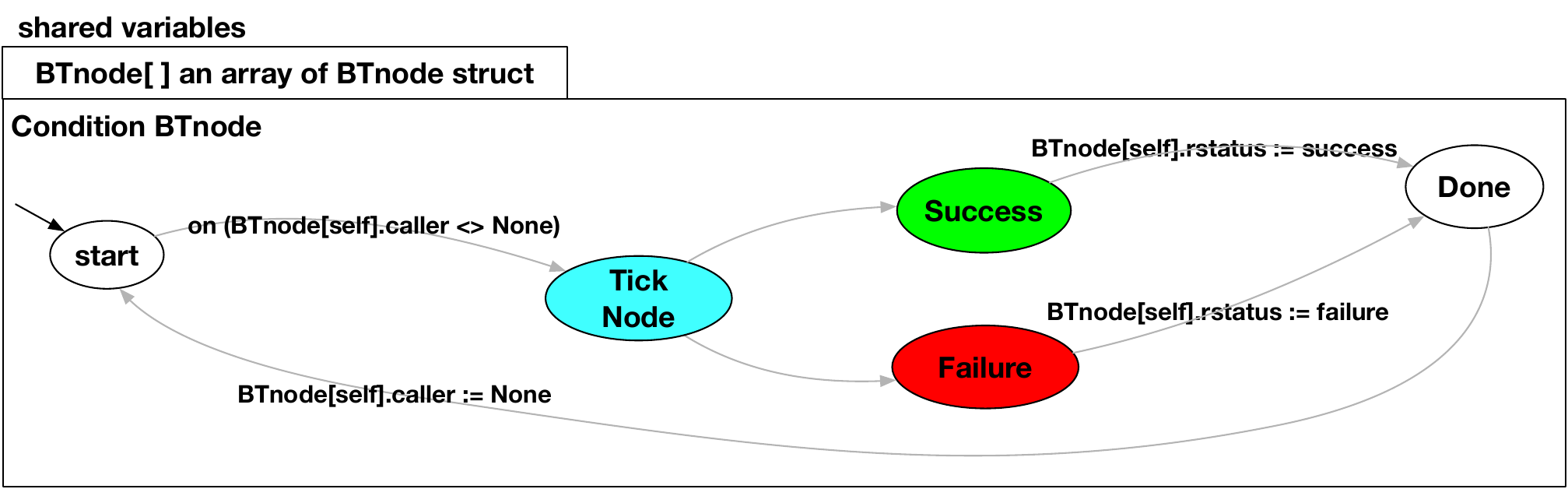}
\caption{The \fiacre{} process modeling the \btn{Condition} node for the verification (\tina{}) model.}
\label{fig:conditiont}
\end{center}
\end{figure*}

\begin{figure*}[!ht]
\begin{center}
\includegraphics[width=0.7\textwidth]{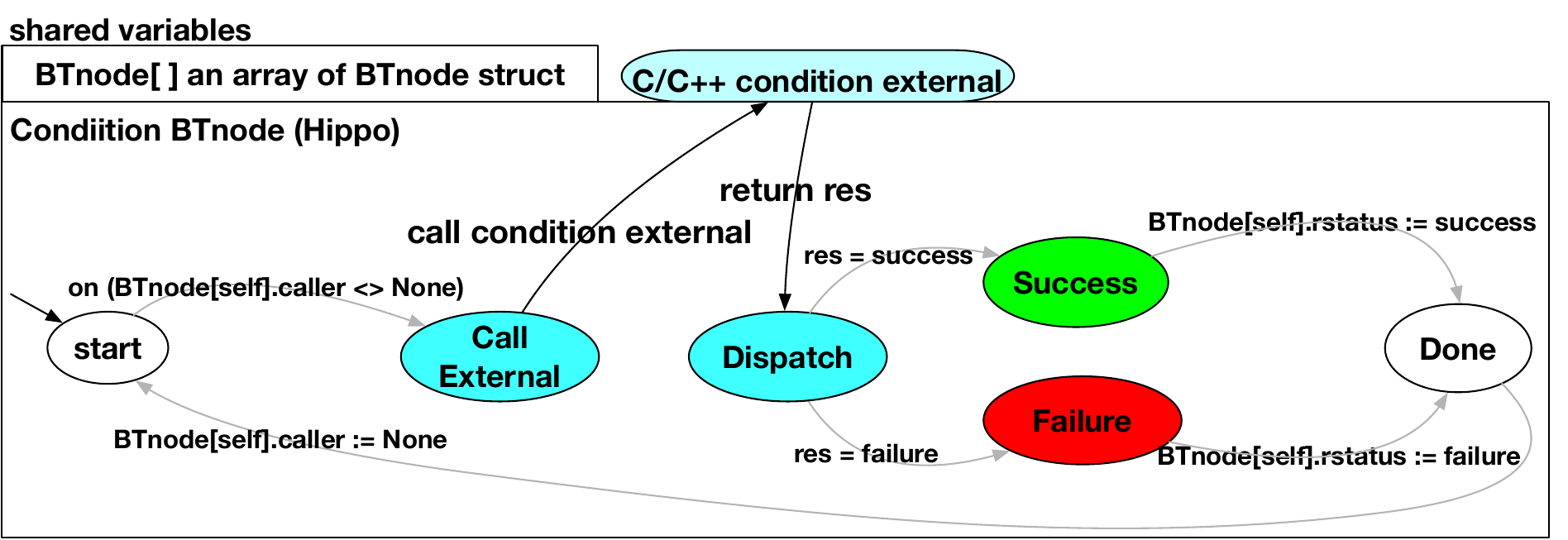}
\caption{The \fiacre{} process modeling the \btn{Condition} node for the runtime (\hippo{}) model.}
\label{fig:conditionh}
\end{center}
\end{figure*}

The \btn{Condition} node is the simplest node, it only returns \success{} or \failure. The \tina{} version will just return either values
(See Figure~\ref{fig:conditiont}\code{p}\pageref{fig:conditiont}), while the \hippo{} version will call the \emph{external} C/C++ function
which performs the test and return its value (See Figure~\ref{fig:conditionh}\code{p}\pageref{fig:conditionh}). The C/C++ function is
expected to be fast and should not delay execution unnecessarily. Note that in our formalism, one can pass arguments to the C/C++ call. This
is very much welcome to make the BT language more expressive and to some extent to have these values available in the formal model.

\subsection{\btn{Action} Node}
\label{sec:btnaction}

\begin{figure*}[!ht]
\begin{center}
\includegraphics[width=0.7\textwidth]{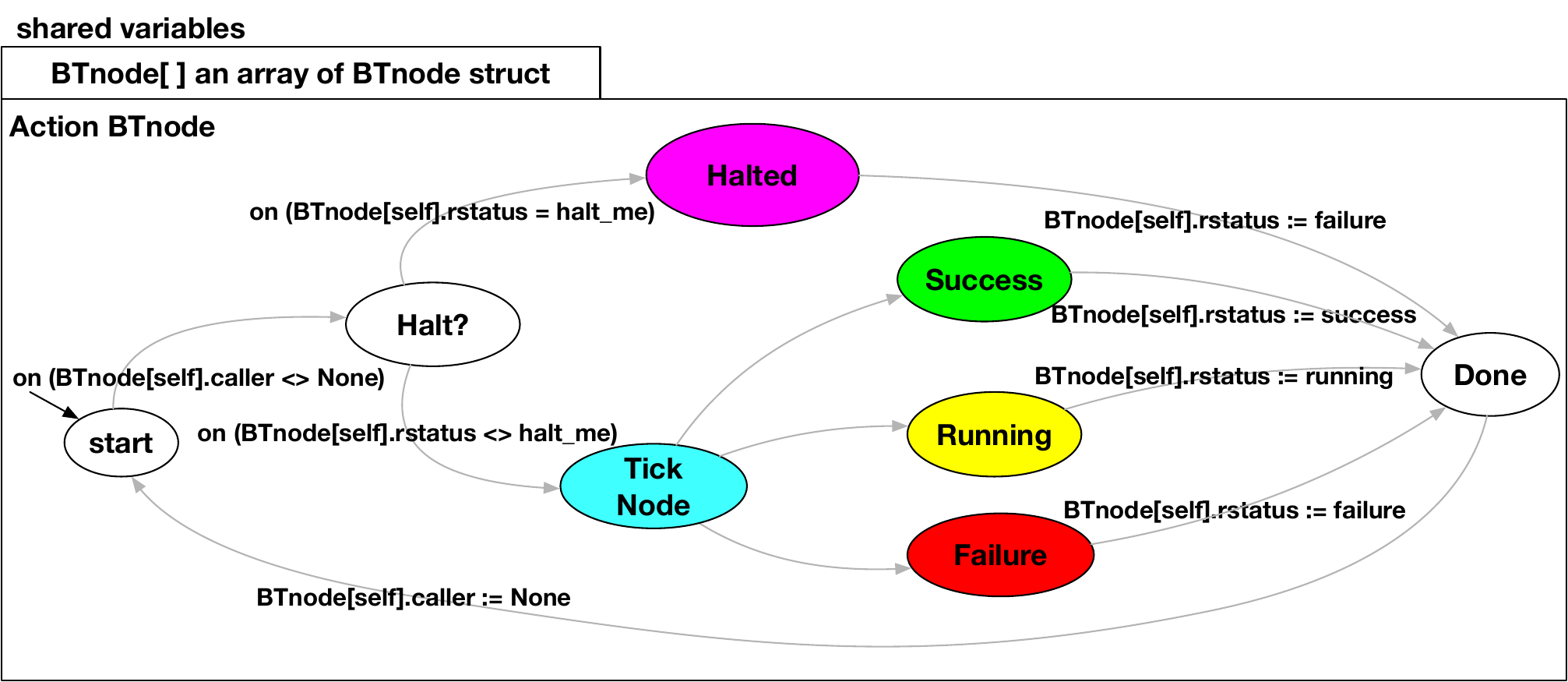}
\caption{The \fiacre{} process modeling the \btn{Action} node for the verification (\tina{}) model.}
\label{fig:actiont}
\end{center}
\end{figure*}

\begin{figure*}[!ht]
\begin{center}
\includegraphics[width=0.7\textwidth]{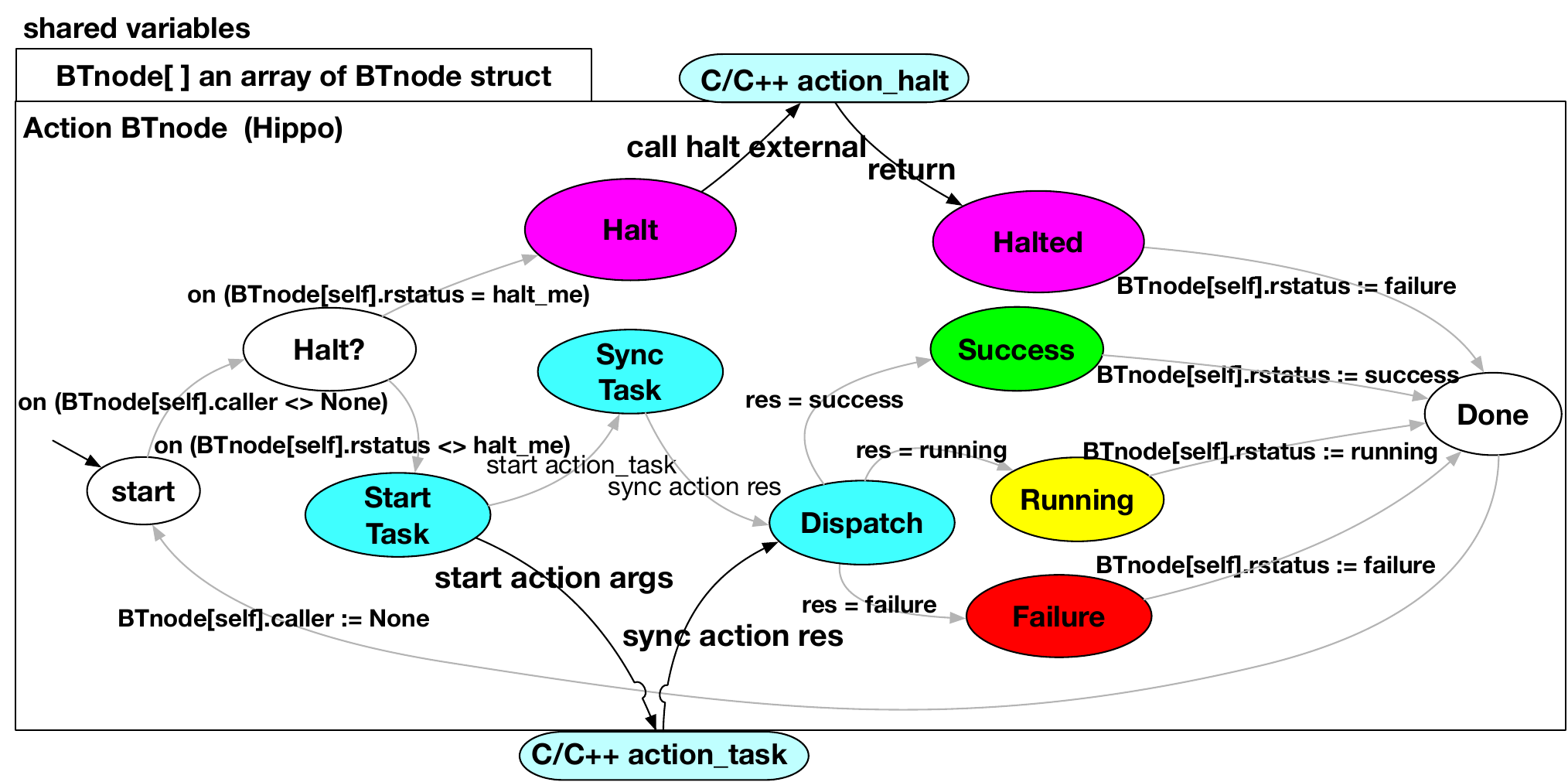}
\caption{The \fiacre{} process modeling the \btn{Action} node for the runtime (\hippo{}) model.}
\label{fig:actionh}
\end{center}
\end{figure*}

The \btn{Action} node is slightly more complex, as it can also return \running{}. Indeed, actions may take some time and not return
\success{} or \failure{} right away. To preserve the reactivity of the overall execution, and keep the tick short, it is often advised to
return \running{} while the action is still executing in its own thread. As a consequence, it means that an \bt{Action} node may be \emph{halted}
(i.e. at some point, it is in a \running{} state but its parent node wants to halt it). In the \tina{} version (See
Figure~\ref{fig:actiont}\code{p}\pageref{fig:actiont}), all the return values are possible, while in the \hippo{} version (See Figure~\ref{fig:actionh}\code{p}\pageref{fig:actionh}), a \fiacre{}
\emph{task} is started calling the C/C++ \code{action\_task}, and subsequent ticks will return \running{} until the C/C++ \code{action\_task} finishes
and returns \success{} or \failure{}. An \emph{external} C/C++ \code{action\_halt} is also defined and is called if the action must be
halted (it then returns \failure{}). While the \code{action\_task} may take some time in its own thread, the \code{action\_halt} is expected to
be fast. In an \btn{Action} node too, one may define and pass some arguments to the C/C++ function.  Listing~\ref{lst:factionh}\code{p}\pageref{lst:factionh} in
Appendix~\ref{app:app1} shows the \fiacre{} code for the \btn{takeoff} action.

Note that only the \btn{Action} and \btn{Condition} nodes have a slightly different \tina{} (offline) and \hippo{} (online) versions. For all the other BT
nodes, the model is strictly the same.

\subsection{\btn{Sequence} Nodes}

The \btn{Sequence} node ticks/executes its child in sequence as they succeed, until one fails. It then returns \failure{}. If all succeed, it
returns \success{}. If one child returns \running{},  the \btn{Sequence} returns \running{}. Upon return of the tick this particular child is ticked/called again.

\begin{figure*}[!ht]
\begin{center}
\includegraphics[width=1\textwidth]{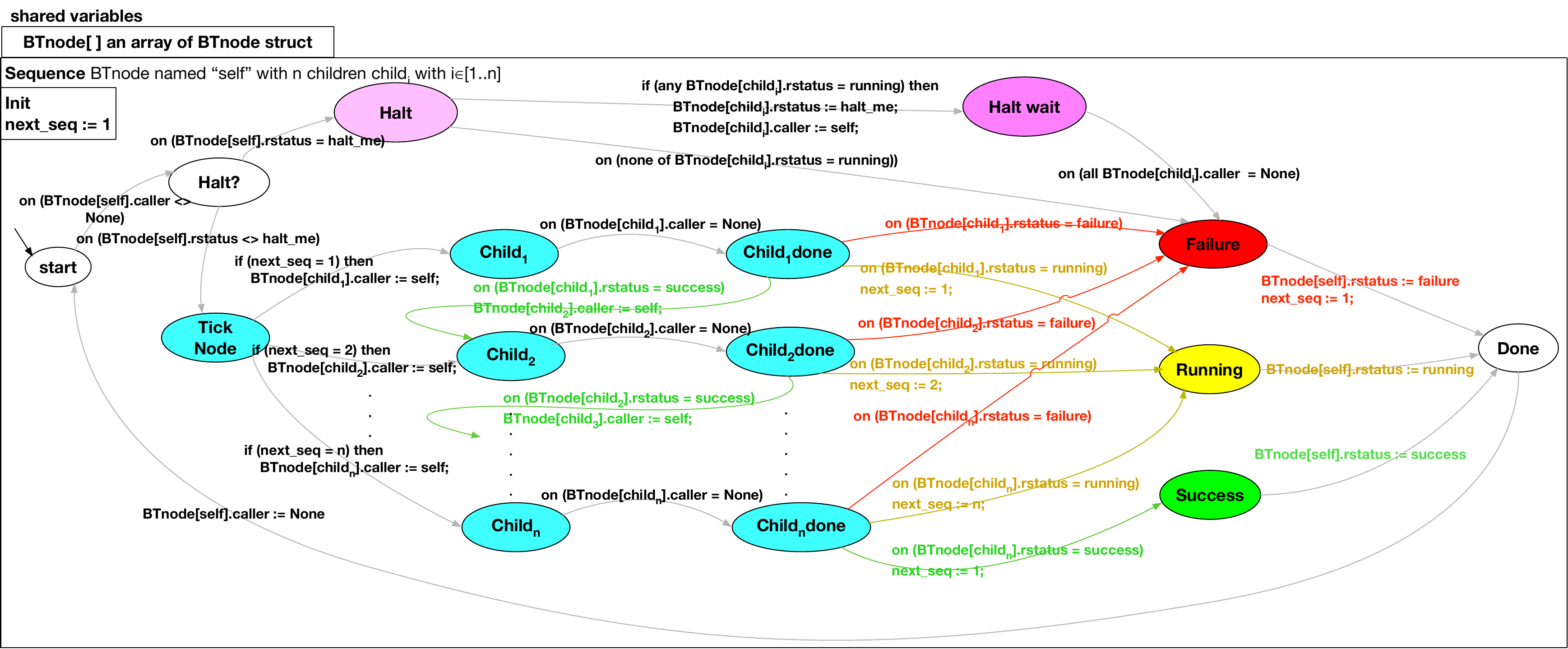}
\caption{The \fiacre{} process modeling the \btn{Sequence} node.}
\label{fig:sequence}
\end{center}
\end{figure*}

Listing~\ref{lst:fsequenceh}\code{p}\pageref{lst:fsequenceh}, shows the complete \fiacre{} code of a \btn{Sequence} node with three children. The \fiacre{} process has a local variable
\code{next\_seq} initialized at 1, which holds which node will be ticked/called when the current node is ticked/called again. From the
\state{Tick Node} state (See Figure~\ref{fig:sequence}\code{p}\pageref{fig:sequence}), one proceeds to the \state{$Child_{\texttt{next\_seq}}$} state, which ticks/calls the proper
child. We wait until this child is done \code{caller=None} and check its returned status \code{rstatus}:

\begin{itemize}
\item \success{}, if it is the last child to succeed, we return \success{}, otherwise, we proceed to the next child
  $Child_{\texttt{next\_seq+1}}$.

\item \failure{}, we return \failure{}.

\item \running{}, we set \code{next\_seq} to the proper value and return \running{}.

\end{itemize}

There exist variants of \btn{Sequence:} \btn{ReactiveSequence}, \btn{SequenceWithMemory}. They alter the way the sequence is ticked after
\failure{} or \running{} are returned. For example \btn{ReactiveSequence} always restarts the sequence from the beginning after one child
has returned \running{}. The goal here is not to list all the variants, just to give a hint on how these are transformed in \fiacre{}. Of
course, the generated \fiacre{} code implements the proper semantics for each variant by properly setting the \code{next\_seq} value.

\subsection{\btn{Fallback} Nodes}

The \btn{Fallback} node executes/ticks its child in sequence as they fail, until one succeeds. It then returns \success{}. If all fail, it
returns \failure{}. If one child returns \running{}, it returns \running{}. Upon return of the tick this child is again ticked/called. As
shown on the Figure~\ref{fig:fallback}\code{p}\pageref{fig:fallback}, it is the symmetric of the \btn{Sequence} node, thus its behavior is just the mirror of the
\btn{Sequence} one described above. There is a variant of \btn{Fallback}: \btn{ReactiveFallback}. It alters the way the \btn{Fallback} is
restarted/reticked after \running{} is returned. \btn{ReactiveFallback} always restarts from the beginning after one child has returned
\running{}. This is achieved in \fiacre{} by properly setting the \code{next\_fb} value.

\subsection{\btn{Parallel} Nodes}

The \btn{Parallel} node specifies how many children \code{m} out of all \code{n} children must succeed for the \btn{Parallel} node to
succeed (with \btn{ParallelAll} they must all succeed: \code{n = m}). From this, we see that the implementation just ticks all
the node, and then keep track of how many return \failure{}, \success{} or \running{} (See Figure~\ref{fig:parallel}\code{p}\pageref{fig:parallel}). If any final
condition for \success{} or \failure{} is met (enough \code{BTnode[child1].rstatus = \success{}})\footnote{(enough... \success{}) is true
  when \code{\success{} >= m}} or (enough \code{BTnode[childn].rstatus = \failure{}})\footnote{(enough...  \failure{}) is true when
  \code{\failure{} > n - m}}, then it proceeds to the corresponding state, otherwise, this node
returns \running{}.  Note that \btn{Parallel} nodes can lead to children being halted (if the condition for \success{} or \failure{} are met
while some children are still \running{}).

\subsection{\btn{Decorator} Nodes}

\begin{figure*}[!ht]
\begin{center}
\includegraphics[width=1\textwidth]{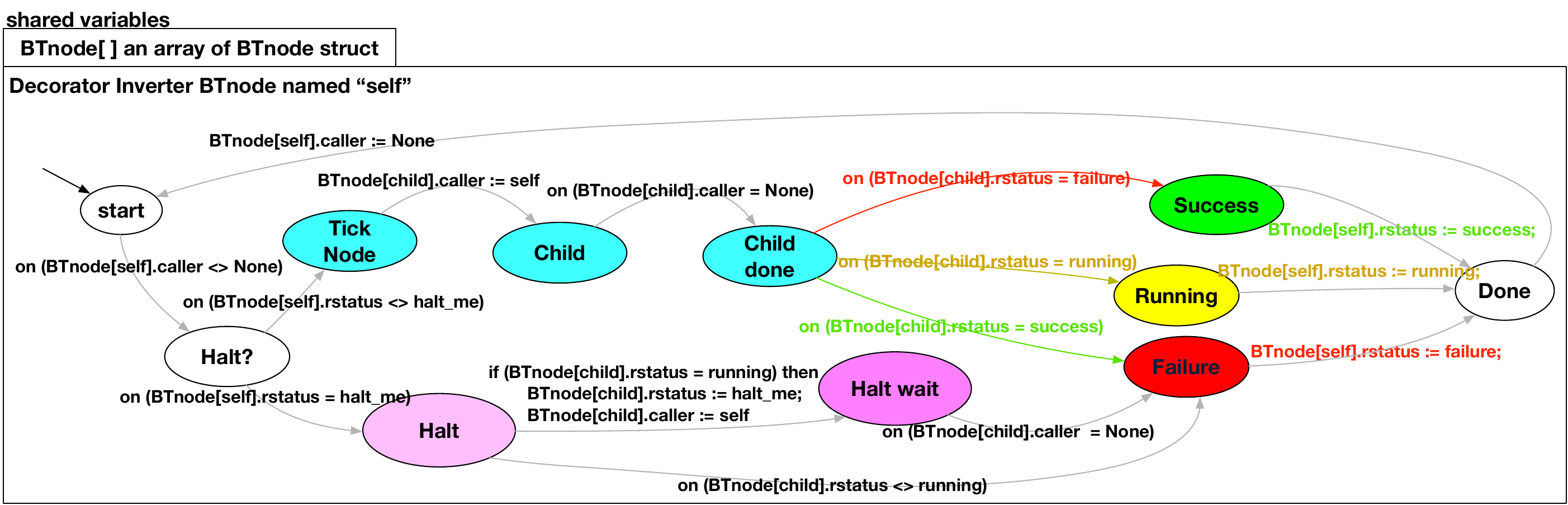}
\caption{The \fiacre{} process modeling a \btn{Decorator} (\btn{Inverter}) node.}
\label{fig:decorator}
\end{center}
\end{figure*}

\btn{Decorator} nodes have only one child. When this child is \state{done} the transitions to \success{}, \failure{} or \running{} depend on
the particular \btn{decorator} type (\btn{Inverter}, \btn{ForceFailure}, \btn{ForceSuccess}, \btn{KeepRunningUntilFailure},
\btn{RetryUntilSuccessful}, \btn{RateController}, \btn{Repeat}, etc). For example, the \btn{Inverter} (See Figure~\ref{fig:decorator}\code{p}\pageref{fig:decorator}) one
will swap the \success{} and \failure{} transitions, while leaving untouched the \running{} one, \btn{Repeat} will call its child a number
of times, etc.

\subsection{Semantics clarification and added Features}

As mentioned earlier, the translation to \fiacre{} defines a formal semantics of BT. But there are particular choices which need to be
clarified and specified. Many of these choices have already been identified~\cite{Ghiorzi:2024aa}, we try here to settle them with a formal
proposition.

\subsubsection{Halting running BT nodes}

We have seen there are a number of situations where one may have to handle ``orphan'' running branches. They are not really orphans, but
the situation is such that their parents have already decided the \success{} or \failure{}, independently of their own final outcome other than
\running{}.  This happens when a \btn{Parallel} fails (or succeeds) and there are still \running{} children, similarly when a
\btn{ReactiveSequence} fails or a \btn{ReactiveFallback} succeeds, one child may still be \running{}. So for these nodes we added 2
keywords \code{:wait} and \code{:halt} to their specification. If \code{:halt} is true, we explicitly halt the still \running{}
children. Note that this is propagated all the way to leaves nodes, in particular \btn{Action} nodes will explicitly halt whatever they are
doing by calling their \code{action\_halt} C/C++ function (see Section~\ref{sec:btnaction}). If \code{:wait} is true, the node waits until
all the children return something else than \running{}. These choices can be discussed and modified, but at least, the semantics is clarified,
defined and implemented in the formal model.

\subsubsection{Ticks and BT root}
\label{sec:tick}

\fiacre{} supports time, and so does the produced TTS and the \tina{} verification tools. We consider that a BT tick takes one \fiacre{}
unit of time\footnote{In the regular \fiacre{} verification language, time is ``unitless'', but in the \hippo{} engine, we set the ``tick''
  (\qty{100}{ms} in the experiments described in this paper).}. By default, only the BT root generates ticks, one after another,
independently of ``how long'' the tick traversal took. So the default BT model produced in \fiacre{} has just one transition $[1,1]$ in the
root BT, and all other transitions are timeless: $[0,0]$. This is a perfectly ``fine'' assumption if one considers the tick more as an
execution ``token'' traversing the BT in no time, but if one wants to perform more advanced validation and verification, it may be a good
idea to propose different tick semantics. For now we propose two other semantics, one is to have all the \btn{Action} and \btn{Condition} BT
nodes to have a $[1,1]$ transition on their \state{tick\_node} state, and the other one is to have all BT nodes with such $[1,1]$ transition.  Of
course, the chosen semantics among the three possibilities can be specified when producing the \fiacre{}
model.  
Note that from a temporal model checking point of view, the last semantics is preferred, as it minimizes simultaneous transitions
interleaving, hence the branching factor to the reachable states exploration.

The BT root is also responsible for keeping the BT alive. What should it do when its child returns \running{}, \success{} or \failure{}?  Some
implementations keep running on \success{}, others do not. To clarify the model we produce, the root BT keeps \running{} while its child returns
\running{}, and stops when it returns \success{} or \failure{} (e.g., in the drone experiment we present in Section~\ref{sec:uav-bt}, this is
the expected behavior).

\subsubsection{State Variables, Expression Evaluation and Node Status}

BT tend to rely on ``external'' \btn{Actions} and \btn{Conditions} to compute values, tests them, etc. One of our objectives is to get as much as possible of the BT and
its associated \btn{Condition} and \btn{Action} nodes modelled in \fiacre{}. The more we get, the more properties we can prove on the reachable state of the
BT, and the more we control execution at runtime.

Hence we introduce \emph{state variables} which can be used in the BT model and will end up in the \fiacre{} model as \fiacre{} variables.

For now, state variables can hold a natural number or an enumeration. See the example below with \sv{flight\_level} which can take an
\code{int} between 0 and 3, or \sv{battery} which can take three values (\svv{Good}, \svv{Low} and \svv{Critical}) (note that for an
enumeration, one can specify the acceptable transitions from one value to others).

We also added two new leaves nodes to the BT specification:
\begin{description}
\item [\btn{SetSV}] nodes can be used to set a state variable passed with the :SV keyword, by calling the :ID function (defined in
  \fiacre{}). \btn{SetSV} nodes only return \success{}. 
\item[\btn{Eval}] nodes evaluate the condition they hold and return \success{} or \failure{}.
\end{description}

Last, we also make available for evaluation the resulting status of any BT, e.g.  \code{(Eval (= camera\_track.rstatus success))} will return
\success{} if the execution of the \code{camera\_track} \btn{Action} node was a \success{}.

\section{Deploying, Model Checking and Running the \fiacre{} BT}
\label{sec:deploy}

The final \fiacre{} model of the BT is produced by instantiating all the BT nodes \fiacre{} processes and combining them in parallel.

 %\subsection{The \tool{frac} compilation}

 As shown on Figure~\ref{fig:workflow}\code{p}\pageref{fig:workflow} both \fiacre{} models (offline and online) are compiled in their
 respective TTS (i.e., a Time Petri Net with data) with the \code{frac} compiler.

\subsection{Offline  formal verification of  BT}
\label{sec:ofv}
\label{sec:ltlprop}

The \tina{} toolbox used here mostly rely on LTL (SE-LTL state/event LTL~\cite{Chaki:2004aa} to be more accurate). LTL already offers a rich
language and a lot of flexibility to write and prove some logical and temporal properties on the formal BT. Using model checking, one can
either check properties on the fly (reachability of a state), or show the absence of a state (but for this, the whole reachable state set has to
be built).  By default, the \tool{sift} tool computes the reachable states set of the BT. The resulting \code{.ktz} file (binary kripke
transition system) can then be analyzed with default or additional properties with the \code{selt} tools\footnote{The \tina{} toolbox
  provides many tools, we just focus here on \code{selt} and \code{sift}.}.

For each BT node, we define and check some default properties:  can the BT execute and complete,
can it succeed, fail, or return running? Can it be halted?  Most of these properties correspond to either a particular \state{state} in the fiacre
model, or to some explicit value in the \code{btnode[self].rstatus} record. So the properties can be synthesized automatically as follow
(by checking if their negation is reachable, i.e. we try to show that the state cannot be reached, and expect \code{selt} to return a
counter example):

\begin{lstlisting}[language=LTL]
// For the  Action_takeoff btnode
prove absent drone/12/state done 
prove absent drone/12/value (btnode[takeoff].rstatus=success)
prove absent drone/12/value (btnode[takeoff].rstatus=failure) 
prove absent drone/12/state halted 
// the 12 is the index of the Action_takeoff btnode process instance in all the processes 
// combined in parallel to build the drone component
\end{lstlisting}

One can also check more complex properties. For example, if we reach a particular state, then we will eventually reach another one. 

We also modelled some examples found in some papers presented in Section~\ref{sec:soa}.

In \cite{Biggar:2020aa} the authors present an example of a Mars rover using unfolded solar panels to charge its battery, but should fold
them when there is a storm. The same example is studied in ~\cite{Wang:2024aa} with the BT transformed to BIP on which they  check the
same properties. The equivalent BT in the \code{.btf} formalism can be found in Appendix~\ref{app:mars}, Listing~\ref{lst:mars}\code{p}\pageref{lst:mars}. After
translation to \fiacre{}, we can build the set of reachable states:
\begin{lstlisting}[language=bash]
sift -stats mars_rover.tts -rsd mars_rover.tts/mars_rover.ktz  
20783 classe(s), 20275 marking(s), 72 domain(s), 103111 transition(s)
0.520s
\end{lstlisting}
and to prove the property  (check that the robot cannot be hit by a storm while its  solar panels are unfolded):
\begin{lstlisting}[language=LTL]
prove absent (mars_rover/3/state Unfolded and mars_rover/1/state Storm)
\end{lstlisting}
we get (as they do):
\begin{lstlisting}[language=fiacre]
never (sv__panel__automata_1_sUnfolded /\ sv__meteo__automata_1_sStorm) 
FALSE
\end{lstlisting}

\code{selt} finds a counter example, i.e., a state where this can happen (which is a problem and need to be fixed).

Similarly, in~\cite{Wang:2024aa} they develop an example with a train having a non null speed while a door is open, we modelled it and reach the same 
formal proof results.

\subsection{Online (or runtime) verification}

The formal executable model is obtained by producing the TTS with the \code{frac} compiler, and then linking the result with the \hippo{}
engine library and the C/C++ code implementing the \btn{Action} and \btn{Condition} \fiacre{} tasks and \fiacre{} externals
(Section~\ref{sec:btnaction} and~\ref{sec:btncondition}). In most case these C/C++ code use the client library which allows to call actions,
services or check topics on ROS nodes~\cite{Quigley:2009tg}; or to make requests to or read ports from \GenoM{} modules~\cite{Dal-Zilio:2023aa}.

As mentioned earlier, our translation from \code{.btf} BT to \fiacre{} defines the formal semantics of the language. We have seen above that we
can prove properties on this formal model. But executing the formal model with \hippo{} is also a way to ``validate'' that the formal
semantics we defined is correct with respect to the operational semantics one expects from BT. In other words, the execution of the formal
model produces what is expected by the original BT programmer, as if he/she were using a regular BT execution engine.

As seen on Figures~\ref{fig:screen-bt}\code{p}\pageref{fig:screen-bt}
and~\ref{fig:archi-uav}\code{p}\pageref{fig:archi-uav}, %and~\ref{fig:screen}\code{p}\pageref{fig:screen},
our BT show changing color while executing: white, the node has not yet been visited; dark blue, the tick is currently in this node; light
blue, the tick has been passed to child(ren) node(s) below and the node is waiting for the children to return; yellow, the node has returned
\running{}; green, the node has returned \success{}; red, the node has returned \failure{}; purple, the node was running and has been halted
by one of its parent nodes; pink, the node has been instructed to halt itself (and its running branches).
  
The white square node close to the root of the BT indicates the \hippo{} tick %, which is set at \qty{10}{Hz} in this experiment, and
which is also the BT tick.

\section{Some illustrating examples}
\label{sec:examples}

We now illustrate our approach with two examples, a drone controller and the Nav2 navigation ROS2 stack.

\subsection{An UAV surveying an area}
\label{sec:uav-bt}

\begin{figure*}[!ht]
\begin{center}
\centerline{\includegraphics[width=1.1\textwidth]{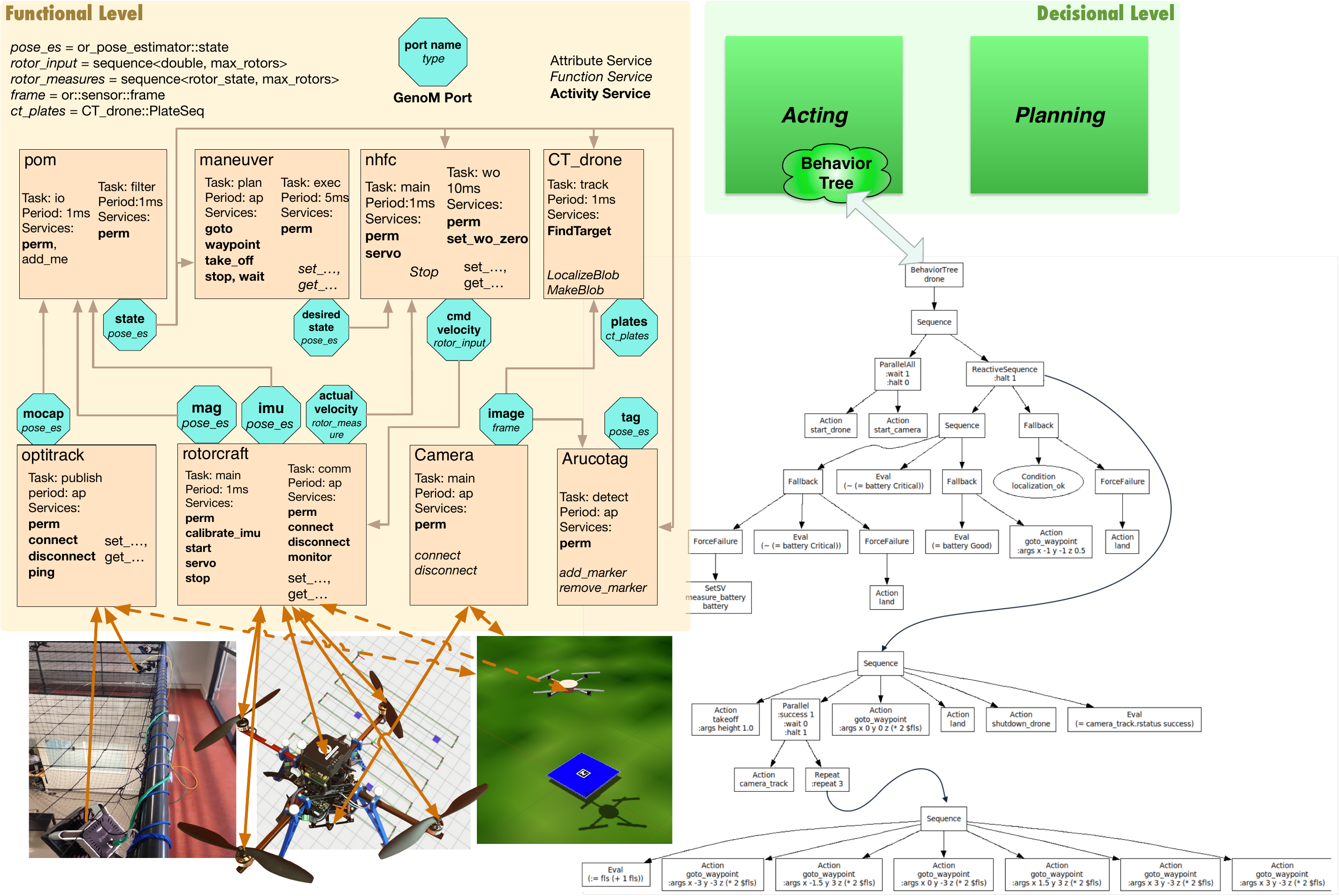}}
\caption{Architecture of the drone experiment. On the left, all the functional components involved, on the right a Behavior Tree in charge
  of the overall mission.}
\label{fig:archi-uav}
\end{center}
\end{figure*}

We program an UAV to perform a survey mission with a BT. The functional layer of this experiment (Figure~\ref{fig:archi-uav}\code{p}\pageref{fig:archi-uav}) has already
been presented in~\cite{Dal-Zilio:2023aa}, but suffices to say that it provides robust localization, navigation, flight control and allows us
to command the drone and its camera. It is deployed using the \GenoM{} specification language (which also maps in the same  formal framework to validate and verify
the functional components~\cite{Dal-Zilio:2023aa}).

Eight functional components are deployed for this experiment. The set of primitive commands available for the \emph{acting} component are:
\bt{start\_drone}, \bt{start\_camera}, \bt{shutdown\_drone}, \bt{takeoff}, \bt{land}, \bt{goto\_waypoint}, and \bt{camera\_survey}. Each of
them has a corresponding \btn{Action} node in the larger BT \bt{drone} on listing~\ref{lst:drone3bt}\code{p}\pageref{lst:drone3bt} (e.g,
\bt{takeoff} (line~\ref{ll:takeoff}), \bt{goto\_waypoint} (lines~\ref{ll:gw1}-\ref{ll:gw2}), etc),

The state variables handled by the model are \sv{battery}: \svv{Good}, \svv{Low} and \svv{Critical} (line~\ref{ll:battery}); and
\sv{flight\_level}: a natural number between 0 and 3 (line~\ref{ll:fls}).

\begin{lstlisting}[caption={A more complex drone survey  BT in the \code{.btf} format, with \emph{state variables} declaration, \btn{SetSV}
and \btn{Eval} nodes. The $:=$ is the assignment operator, and $\sim$ is the logical negation (See also Figure~\ref{fig:bt-drone3}\code{p}\pageref{fig:bt-drone3}).}, numbers=left, xleftmargin=15pt, label={lst:drone3bt}, language={[btf]Lisp}]
((defsv fls (*@ \label{ll:fls} @*) ; the flight level of the drone
  :init 0
  :min 0
  :max 3)

(defsv battery (*@ \label{ll:battery} @*) ; the battery level of the drone
  :states (Good Low Critical)        ;Self explanatory
  :init Good
  :transitions :all)

(BehaviorTree :name drone
    (Sequence
      (ParallelAll :wait 1 :halt 0 (*@ \label{ll:wait} @*) 
        (Action :ID start_drone)
        (Action :ID start_camera))
      (ReactiveSequence :halt 1 (*@ \label{ll:halt} @*) 
        (Sequence
          (Fallback
            (ForceFailure :ID fail
              (SetSV :ID measure_battery :sv battery)) (*@ \label{ll:setsv} @*) 
            (Eval (~ (= battery critical))) (*@ \label{ll:eval} @*) 
            (ForceFailure :ID fail_mission
              (Action :ID land)))
          (Eval (~ (= battery critical)))
          (Fallback
            (Eval (= battery good))
            (Action :ID goto_waypoint :args (x -1 y -1 z 0.5)))) ; goto charging station
        (Fallback
          (Condition :ID localization_ok)
          (ForceFailure :ID fail
            (Action :ID land)))
        (Sequence
          (Action :ID takeoff :args (height 1.0 duration 0)) (*@ \label{ll:takeoff} @*)
          (Parallel :success 1 :wait 0 :halt 1 ; if the tracking or the nav success... we are done.
            (Action :ID camera_tracking :name camera_track)
            (Repeat :repeat 3 (*@ \label{ll:repeat} @*) 
              (Sequence
                (Eval (:= fls (+ 1 fls))) (*@ \label{ll:add1} @*) 
                (Action :ID goto_waypoint :args (x -3 y -3 z (* 2 $fls)))  (*@ \label{ll:gw1} @*) 
                (Action :ID goto_waypoint :args (x -1.5 y 3 z (* 2 $fls)))
                (Action :ID goto_waypoint :args (x 0 y -3 z (* 2 $fls)))
                (Action :ID goto_waypoint :args (x 1.5 y 3 z (* 2 $fls)))
                (Action :ID goto_waypoint :args (x 3 y -3 z (* 2 $fls)))
                (Action :ID goto_waypoint :args (x 3 y -3 z (* 2 $fls))))))
          (Action :ID goto_waypoint :args (x 0 y 0 z 5))  (*@ \label{ll:gw2} @*) 
          (Action :ID land)
          (Action :ID shutdown_drone)
          (Eval (= camera_track.rstatus success))))))) (*@ \label{ll:finaleval} @*) 
\end{lstlisting}

\begin{figure*}[!ht]
\begin{center}
\includegraphics[width=0.97\textwidth]{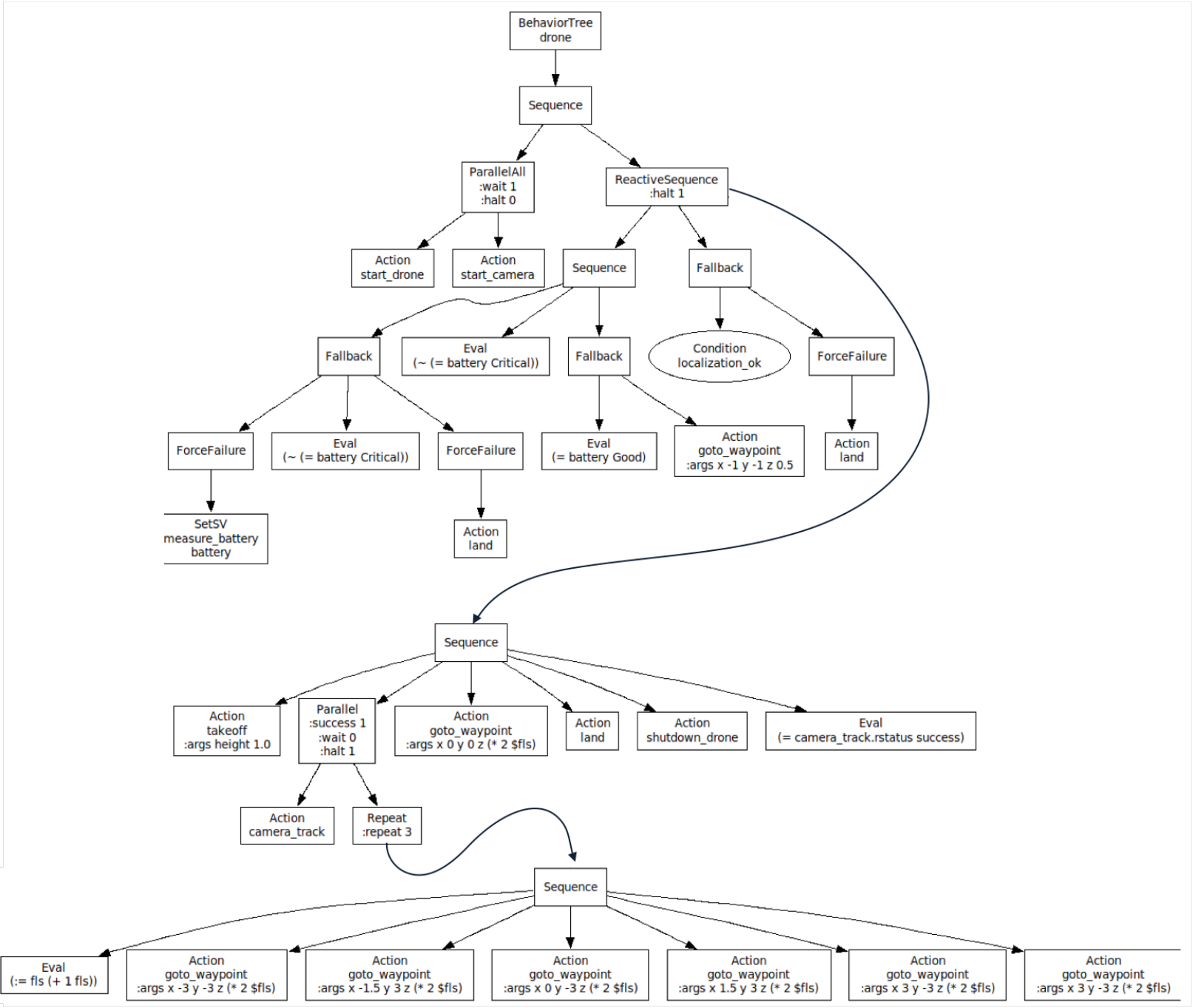}
\caption{The graphical representation of the BT Listing~\ref{lst:drone3bt}\code{p}\pageref{lst:drone3bt}.}
\label{fig:bt-drone3}
\end{center}
\end{figure*}
 
Calling \bttf{} results in a \fiacre{} model with two processes for the state variables, and thirty eight processes for the BT
nodes.\footnote{The resulting code can be found in the \code{examples} subdirectory of
  \url{https://redmine.laas.fr/projects/bt2fiacre/repository}.}

\subsubsection{Offline verification results}
\label{sec:results}
\label{sec:cpu}

To compute the reachable states set of the drone BT (Listing~\ref{lst:drone3bt}\code{p}\pageref{lst:drone3bt}) we call the  \code{sift} tool:

\begin{lstlisting}[language=bash]
sift -stats drone.tts drone.tts/drone.ktz
49 196 302 classe(s), 49 145 735 marking(s), 152 domain(s), 267 702 880 transition(s)
4552.704s
\end{lstlisting}

which takes around 1 hour and 16 minutes to build on an Intel(R) Xeon(R) Silver 4110 CPU @ 2.10 GHz with 256 GB of memory.  The resulting
\code{.ktz} has $49 196 302$ classes, $49 145 735$ markings, $152$ domains, $267 702 880$ transitions.  In this context, a marking is a
particular set of states and values for all the processes and variables in the system. A class is a state extended with timing information
on the enabled transitions (therefore we can have several classes with the same marking).

All the default properties presented in Section~\ref{sec:ltlprop} have been checked and all the results are the expected ones. Most of them
take between less than a second and 300 seconds to be verified. As expected, the ones for which the model checker finds a counter example
are usually quick to be verified while the ones which require to explore the whole set take more time.

Among the default proven properties, some are still interesting to comment on. For example, despite the \btn{ParallelAll} on the
\btn{start\_drone} and the \btn{start\_camera}, they cannot be halted (because the  \btn{ParallelAll}  has \code{:halt 0}). Unlike the BT
children of the \btn{ReactiveSequence} which has \code{:halt 1}, they can all be halted (except the first branch)\footnote{As only a branch returning
\running{} on the previous tick may be halted if one branch on its left fails, hence, never the first branch.}.

But on the top of the default properties, we can add new ones specific to this particular experiment. Can we prove that if the \sv{battery}
is \svv{Critical} then the drone will attempt to land, and if it is \svv{Low} it will return to the charging station?  Can we show that if
the camera survey fails then the whole drone mission will fail? Can the drone fly higher than 6 meters?

These properties can easily be written in \fiacre{} pointing to states of BT \fiacre{} nodes (e.g., \state{tick\_node}, \state{failure}),
values of state variable (e.g. \sv{fls}, \sv{battery}), or even returned status from executed BT (e.g.,
\code{btnode[localization\_ok\_btn18].rstatus = failure})\footnote{The $\Box,\Rightarrow,\diamondsuit$ operators have the usual LTL
  semantics.}:

\begin{lstlisting}[language=LTL]
property attempt_to_land_if_battery_Critical is
ltl ([] ((drone/5/value (battery=Critical)) and (drone/5/state done)) =>
    	<> (drone/8/state tick_node)) // land_btn12

property attempt_to_go_to_charging_station_if_battery_Low is
ltl ([] ((drone/12/value (battery=Low))  and (drone/5/state done)) =>
    	<> (drone/13/state tick_node)) // goto_waypoint_btn16

property attempt_to_land_if_localization_failure is
ltl ([] ((drone/16/value (btnode[localization_ok_btn18].rstatus = failure) and (drone/16/state done))) =>
    	<> (drone/17/state tick_node)) // land_btn20

property camera_fails_implies_mission_fails is
ltl ([] (drone/21/state failure) => 	    // camera_tracking_camera_track 
    	<> (drone/39/state failure))        // BehaviorTree1_drone

property fly_not_higher_than_6m is absent (drone/35/value (fls > 3))
\end{lstlisting}

The result are proven true for all these properties:

\begin{lstlisting}[language=fiacre]
operator attempt_to_land_if_battery_Critical : prop
TRUE
0.001s
operator attempt_to_go_to_charging_station_if_battery_Low : prop
TRUE
0.001s
operator attempt_to_land_if_localization_failure : prop
TRUE
0.001s
operator camera_fails_implies_mission_fails : prop
TRUE
0.001s
operator fly_not_higher_than_6m : prop
TRUE
242.462s
\end{lstlisting}

More interestingly, we can also take advantage of \fiacre{}/\tina{} timed models. For example, if one considers the one tick per node semantics
(See Section~\ref{sec:tick}), we can prove that the a critical battery will always leads to a landing within a $[0,2]$ ticks interval
(everything else considered in the BT):

\begin{lstlisting}[language=LTL]
property attempt_to_land_if_battery_Critical_timed is
((drone/5/value (battery=Critical))and (drone/5/state done)) leadsto (drone/8/state tick_node) within [0,2]
\end{lstlisting}

which results in: 

\begin{lstlisting}[language=fiacre]
operator attempt_to_land_if_battery_Critical_timed : prop
TRUE
49.170s
\end{lstlisting}

\subsubsection{Runtime verification results}
\label{sec:hresults}

The \bt{Drone} BT execution by \hippo{} runs as expected, the survey mission is executed, and by adding random fault (on \sv{battery} level,
or failing the \btn{localization\_ok} \btn{Condition} node), we show that the drone behaves as expected (perform land to prevent further
problems). If the survey finds the object, it returns \success{} and the navigation is halted (because the \btn{Parallel} is \code{:success 1
  :wait 0 :halt 1}. Note that at the end, we check with an \btn{Eval} mode, that the returned status of the \code{camera\_track}
\btn{Action} node is \success{}. So when the overall \btn{Drone} BT completes, it will return \success{} if and only if the target was found,
\failure{} otherwise.

Again, the fact that the BT formal model execution exhibit what is expected from the original programmer is a good sign that the formal
semantics is consistent with the operational one, and that the offline proofs we did on the very same model hold for the original
BT.\footnote{Some videos of these runs and the BT animation are available here:
  \url{https://redmine.laas.fr/projects/bt2fiacre/pages/index\#yet-another-more-complex-example-drone3}.}

From a performance point of view, the overhead of the \hippo{} execution is negligible. After all, the \hippo{} engine is a Petri Net (TTS)
``execution'' engine, and has 497 places and 1113 transitions and manages up to 7 task execution threads (one for each \btn{Action} BT node).

\begin{figure*}[!ht]
\begin{center}
\centerline{\includegraphics[width=1.1\textwidth]{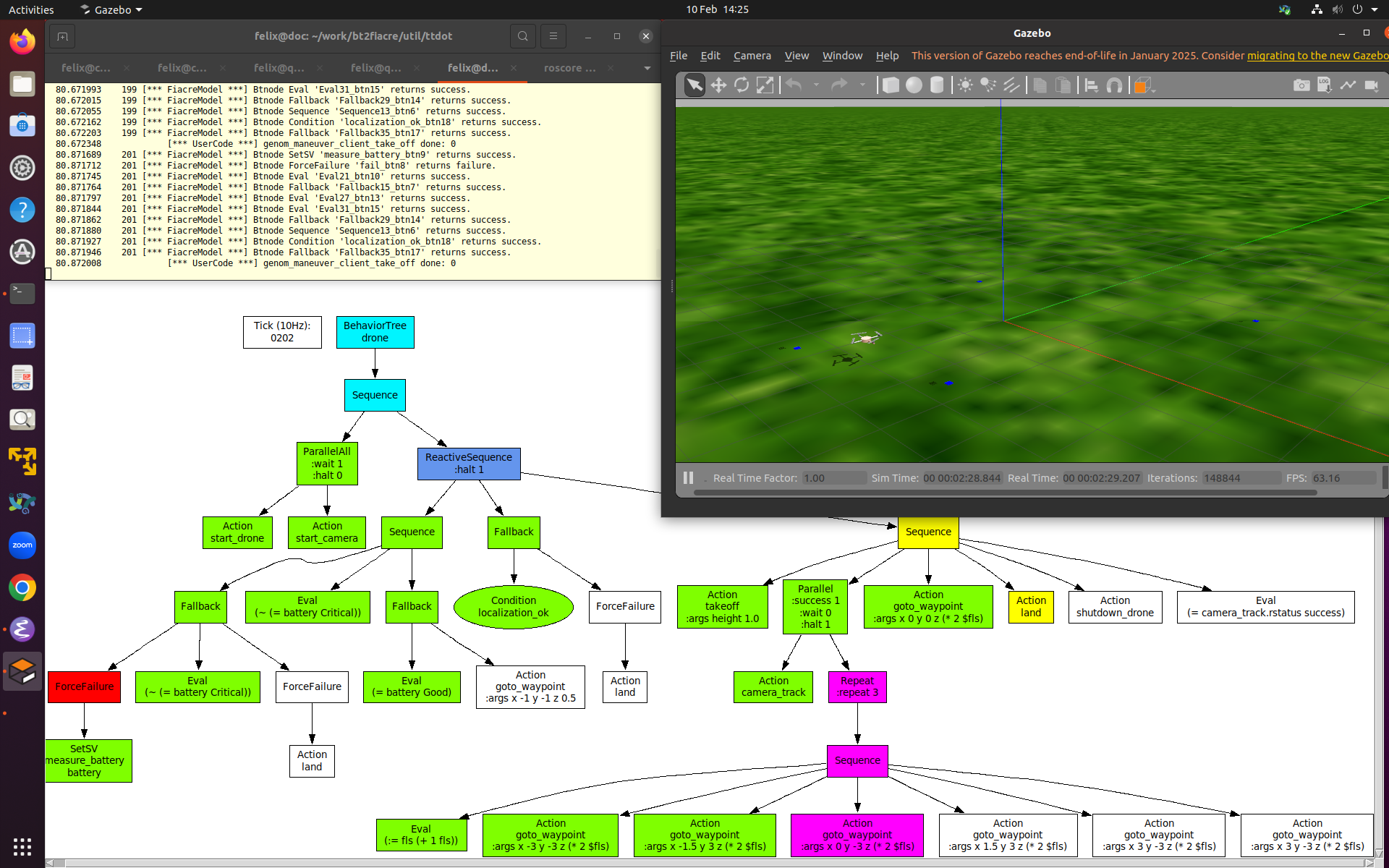}}
\caption{The screen dump of the drone BT mission~\ref{lst:drone3bt}\code{p}\pageref{lst:drone3bt} executing in \hippo{}. This shows that the top \btn{BehaviorTree} and
  its child \btn{Sequence} child have passed the tick to the \btn{ReactiveSequence} currently executing while its \btn{Sequence} has
  returned \running{} from its \btn{Land} child, while the \btn{camera\_track} has returned \success{}, as a result, the \btn{Parallel} has
  halted the \btn{Repeat} and its children. Note the tick counter near the root which indicates the mission is in its 20th second of execution.}
\label{fig:screen-bt-drone}
\end{center}
\end{figure*}

\subsection{ROS2 Nav2 BT}
\label{sec:nav2}

One of the most popular navigation stacks in robotics is the Nav2 one. It is now part of ROS2 and is proposed with a number of BT to
implement it~\cite{ros2nav2:2024aa}.  So to also properly test our approach we implemented the Nav2 BT in our framework.  Note that these BT
define some new \btn{Decorator} and \btn{Control} nodes: \btn{Recovery}, \btn{PipelineSequence}, \btn{RateController}, \btn{RoundRobin},
etc.  They were implemented in our \bttf{} tool and are being  translated to \fiacre{} along the regular ones. One interesting aspect of our approach is that the
operational semantics of these new nodes can be proven in our formal semantic of \fiacre{} using the \tina{} toolbox. For example, one can
formally prove that our implementation of \btn{Recovery} is correct by showing that the second node can only be called when the first one
has returned \failure{}, and that the first one can only be called again when the second has returned \success{}, etc.

% We were able to produce the  reachable states of some of the Nav2 BT, but others remain difficult to synthesize, mostly due to the
% \btn{RateController} nodes. Indeed, this node is implemented with a natural number which counts the number of ticks required to reach the
% tick rate divided by the controller rate. As a consequence, this multiply the number of reachable states of the BT by the maximum value of
% this counter (10, with a \qty{10}{\Hz} tick).

We are able to check the reachable state of the Nav2 BT presented in Appendix, Listing~\ref{lst:ros2nav2btf}\code{p}\pageref{lst:ros2nav2btf} and Figure~\ref{fig:screen-nav2}\code{p}\pageref{fig:screen-nav2}.

\begin{lstlisting}[language=bash]
sift -stats ros2-nav2.tts ros2-nav2.tts/ros2-nav2.ktz
171 656 098 classe(s), 171 656 098 marking(s), 66 domain(s), 422 970 818 transition(s)
5776.872s

selt ros2-nav2.tts/ros2-nav2.ktz ros2-nav2.tts/ros2-nav2.ltl -b
Selt version 3.8.0 -- 05/02/24 -- LAAS/CNRS
ktz loaded, 171 656 098 states, 422 970 818 transitions
2212.781s
\end{lstlisting}

Building the set of reachable states of this BT takes \qty{1}{\hour} \qty{46}{\minute} on the same CPU as the one used in
Section~\ref{sec:cpu} and results in reachable state sets larger than the one we had with the drone experiment. Similarly, all the default
formal properties were checked (taking between 0 and 85 seconds for each property) without any unexpected results.

More interestingly is to check that our implementation of the added BT (e.g., \btn{Recovery}, \btn{RoundRobin}, etc) is correct.

\paragraph{Recovery}
The \btn{Recovery} node\footnote{\url{https://docs.nav2.org/behavior_trees/overview/nav2_specific_nodes.html}} is a control flow
node with two children. It returns \success{} if and only if the first child returns \success{}. The second child will be executed only if
the first child returns \failure{}. If the second child succeeds, then the first child will be executed again. The user can specify how
many \code{:retry} times the recovery actions should be taken before returning \failure{}.

We define a simple generic example:
\begin{lstlisting}[caption={A simple \btn{Recovery} node.}, numbers=left, xleftmargin=15pt, label={lst:recoverybtf}, language={[btf]Lisp}]
((BehaviorTree :name bt_recovery
     (Recovery :num_retries 1 :name recovery
          (Action :name action)
          (Action :name recov))))
\end{lstlisting}

for which the model checker \tool{selt} is able to prove the following properties:

\begin{lstlisting}[language=LTL]
// Action_action		// 1 
// Action_recov			// 2
// Recovery_recovery		// 3
// BehaviorTree_bt_recovery	// 4

property failure_recov_leads_to_failure is // TRUE if the recovery action fails, than the recovery fails
(bt_recovery/2/state failure) leadsto (bt_recovery/3/state failure) within [0,0]

property failure_action_leads_to_recovery_failure is  // FALSE because retry is still 0 (not retried yet)
(bt_recovery/1/state failure and bt_recovery/3/value (retry = 0))  leadsto  (bt_recovery/3/state failure) within [0,0]

property failure_action_leads_to_recovery_failure is // TRUE because we already retied
(bt_recovery/1/state failure and bt_recovery/3/value (retry = 1))  leadsto  (bt_recovery/3/state failure) within [0,0]

property success_action_leads_to_success is // TRUE
(bt_recovery/1/state success) leadsto (bt_recovery/3/state success) within [0,0]
\end{lstlisting}

We present here a proof on a simple instance, but similar proofs can be made on an instance of \btn{Recovery} embedded in a larger BT, for
example the ones in the Nav2 BT in Appendix, Listing~\ref{lst:ros2nav2btf}\code{p}\pageref{lst:ros2nav2btf} and
Figure~\ref{fig:screen-nav2}\code{p}\pageref{fig:screen-nav2}.

\paragraph{RoundRobin}
The \btn{RoundRobin} node\footnote{\url{https://docs.nav2.org/behavior_trees/overview/nav2_specific_nodes.html}} is a control node which
ticks its children in a round robin fashion until a child returns \success{}, in which case the parent node will also return \success{}. If all
children return \failure{} so will the parent \btn{RoundRobin}.

Similarly, we define a simple generic example:
\begin{lstlisting}[caption={A simple \btn{RoundRobin} node.}, numbers=left, xleftmargin=15pt, label={lst:roundrobinbtf}, language={[btf]Lisp}]
((BehaviorTree :name bt_roundrobin
    (KeepRunningUntilFailure :name kr
      (RoundRobin :name RR
          (Action :name A1)
          (Action :name A2)
          (Action :name A3)
          (Action :name A4)))))
\end{lstlisting}

for which the model checker is able to prove the following properties:

\begin{lstlisting}[language=LTL]
// Action_A1	// 1
// Action_A2	// 2
// Action_A3	// 3
// Action_A4	// 4
// RoundRobin_RR // 5

property failure_a1_leads_to_failure is // FALSE a failure of one child does not necessary leads to rr failure
(bt_roundrobin/1/state failure) leadsto (bt_roundrobin/5/state failure) within [0,1]

property failure_a4_leads_to_a1_ticked is  // TRUE a failure of a child with less than 3 failure so far leads to the next child to be ticked
((bt_roundrobin/4/state failure ) and (bt_roundrobin/5/value (failed < 3))) leadsto (bt_roundrobin/1/state tick_node) within [0,0]

property success_a2_leads_a3_ticked_exp_true is // TRUE success of a child leads to the next one to be ticked upon rr reticked
(bt_roundrobin/2/state success) leadsto (bt_roundrobin/3/state tick_node) within [0,3] // 3 because the tick has to go all the way up

property success_a2_leads_rr_success_exp_true is // TRUE
(bt_roundrobin/2/state success) leadsto (bt_roundrobin/5/state success) within [0,0]
\end{lstlisting}

Similar proofs can be made with any BT nodes to show that our \fiacre{} implementation satisfies the expected formal properties
defining the BT node considered.

\subsubsection{Runtime}

All the Nav2 BT have been translated to \code{.btf} and run with Hippo and simulated ROS2 actions. One of the features we added and we
believe goes beyond the ``nice graphical touch'' aspect, is to dynamically color the BT nodes while they execute. Of course, there are text
traces of the Hippo engine running the BT, but being able to  follow the execution ticks as well as the last returned status for each BT is rather
informative.  Some videos of these runs and the BT animation are available here:
\url{https://redmine.laas.fr/projects/bt2fiacre/pages/index\#playing-with-ros2-nav2-bt}.
Figure~\ref{fig:screen-nav2}\code{p}\pageref{fig:screen-nav2} shows the Nav2 BT (See Listing~\ref{lst:ros2nav2btf}\code{p}\pageref{lst:ros2nav2btf}) executed by \hippo{}.

\section{Limits, discussion, future work, and conclusion}
\label{sec:conclusion}

Before concluding, we propose to examine the limits of our work, to discuss its pros and cons and consider future work.

\subsection{Limits}

Some limits of our work are mostly due to choices currently made, which could be reconsidered, if needed. For example, we do not implement a
black board to handle variables. Instead we propose to handle BT variables in \fiacre{} directly. This has the advantage of being able to
include these variables in the formal model, hence in the proof and the runtime verification (e.g. the \sv{fls} flight level in the drone
experiment). The disadvantage is that only \fiacre{} supported types can be used.

An area where our approach would suffer, is when BT are dynamically modified, or transformed. For example in
Section~\ref{sec:soabtplanning} we consider BT used for planning. In these approaches, BT are being modified on the fly, clearly, we could not
perform model checking on the fly, yet assuming the dynamic of the planner remains slow (below one second) we could still synthesize and
compile the model (both are almost instantaneous), and jump in its execution on the fly. 

Although all our examples are based on one BT, absolutely nothing prevents us from having multiple BT executing together. Yet from a
verification point of view, this would probably lead to an intractable model as models running in parallel tend to multiply the size of their
reachable states set, to account for all the possible transition interleavings.

This brings us to the main limiting factor of our approach. Offline verification with model checking may lead to large intractable reachable
states set. This is a well known limit of these approaches and we do not have any magic bullet. It is hard to predict what will be the size
of the reachable states set for a given BT. Some very large BT may still produce a rather small reachable states set, while a simple BT
(e.g., with a lot of parallelism, loops, etc) are untractable. Yet we have seen in our examples that we can verify reasonably complex
robotics BT skills, that we can prove ``individual'' BT node behavior (e.g. the new control nodes added in Nav2), and that we can also
handle complex properties on missions written with BT. For the models too complex to be verified offline, we can still use the complete online
version with some added monitors, or work on an abstracted model for offline verification (but we would then loose the offline/online
equivalence).

\subsection{Discussion}
\label{sec:discussion}

Our approach to improve the trust one can put in BT completely relies on a on a well established formal language and framework (\fiacre{}
and TTS), a formal verification toolbox (\tina{} to check LTL, CTL, properties, patterns~\cite{Abid:2014aa}\footnote{Patterns allow the
  verification of properties with explicit time (e.g. to prove that at most x units of time will elapse between two states.).}) and
\hippo{}, a runtime engine able to execute the formal
model.  We have seen in Section~\ref{sec:soabtfm} that there are other approaches which transform BT in a formal framework, or harness a
formal model around BT, but to our knowledge, none of them support the \emph{automatic} translation of BT to an equivalent formal model
\emph{and} the execution of the same formal model in place of the original BT. This is a very strong argument as the same formal model,
which clarifies and specifies the formal semantics of BT, can be used to prove properties and also execute and show, while running, that it
properly implements the expected operational semantic, moreover knowing that the proof made offline also holds online.

Despite the formal and proven tools deployed, our translation tool \bttf{} may still contain errors and bugs, but again, the fact that we
can use theses tools to prove the resulting translated model against specifications done on the BT is reassuring.  One could take every
single BT node type, write its formal specification and prove that the \fiacre{} translation satisfies them with \tina{} (as we did for the
ones added in the Nav2 experiment).

% Again, stress
% that having a model both for verification and to run the experiment is a very strong argument in favor of this approach. While writting the
% BT, we can check that it does what its programmer intended it to do, and it can run verification tools on the model.

Furthermore, \fiacre{} offers some features which could be valuable to be added to BT. Explicit time representation is the foremost feature which could
come handy if added to the BT. Indeed, robot planning, acting and control is usually handling explicit time representation. Control loops
have frequency, plans have duration and deadline, actions take some time to execute, etc. So it would be perfectly logical to add explicit
time representation to BT. As a consequence, this would also have an impact on the tick semantic which would probably become more an
execution token than a time elapsing counter.

Another domain where our approach and \fiacre{} could offer some valuable addition is with environment modelling. BT are intrinsically
embedded operational models, they really exist to be executed in the real world (or in simulation). But now that we can consider model
checking them, one needs to take into account (when possible) the outcomes of actions or conditions in the environment. Without any
particular information, the model checker consider all possibilities (\success{}, \failure{} or \running{}), but the real world often prove
to be more ``constrained'', and one could consider a more accurate and explicit model of the environment (e.g., with external asynchronous
events, state variable values changes, etc).

Parallelism is clearly allowed and properly modelled in our framework. Moreover the model checker can verify properties even if it needs to
consider all the possible execution interleavings of parallel nodes (but at some cost). Meanwhile, the semantic of transition in
\fiacre{} is such that resource checking and reservation on a transition are atomic. As a consequence, resources management, even
considering parallelism comes for free with \fiacre{}.

Last, our approach has been implemented in a software suite~\cite{Ingrand:2024ab}  which has been  applied to many BT examples (available
here: \url{https://redmine.laas.fr/projects/bt2fiacre/pages/index}). Moreover, all the examples presented here are available in the
\code{examples} directory and we invite other formal approach to test their systems on these examples and report their results for comparison. 

%\cite{Ingrand:2024aa}

\subsection{Future work}

Considering the tight links between the \fiacre{} framework and our approach/tool \bttf{}, it is clear that any improvement in the former
may also improve the latter. The \fiacre{} developers are considering extending the data types handled in the language by adding rational
numbers and strings. Both types would still allow model checking and enable the deployment of richer state variables in our \code{.btf}
format.

Similarly, there are many existing features in \fiacre{} which could be used in \code{.btf} BT (some have already been used in
\ps{}~\cite{Ingrand:2024aa}, another robotics acting language mapping to \fiacre{}). As already mentioned in Section~\ref{sec:discussion}
above, time would be a valuable addition to model, with time interval $[min,max]$, how long an \btn{Action} is expected to take, or to wait
a given amount of time before returning from a \btn{Wait} node, etc. Not only would this enrich the \code{.btf} format and language, but it
would also be taken into account by the \tina{} model checking tools, and enforced by the \hippo{} engine.

External asynchronous events and state variables asynchronous value changes are features also available in \fiacre{} and its tools. This
again could enrich the \code{.btf} format and allow the programmer to account for ``uncontrollable''  state transitions.

%The \fiacre{} language is getting new basic types, string and rational number. This should open the language to more complex BT model

\subsection{Conclusion}

BT are more and more popular in robotics. To encourage their deployment and improve the trust one has in robotic applications using BT, we
propose an approach and an automatic tool to transform any BT in a formal model with a formal semantics. The resulting models can then be
used offline with model checking to prove some properties of the BT, but also linked to the real actions and perceptions of the robots and
executed online on the robot.  Of course, this can be deployed in BT applications outside of robotics, and also participates to define
extensions to BT and to better formalize them.

\section*{Acknowledgement}

 We thank Bernard Berthomieu, Silvano Dal Zilio and Pierre-Emmanuel Hladik for their help while developing and deploying the work presented here.

\ifHAL
\bibliographystyle{abbrvnat}
\else
\bibliographystyle{elsarticle-harv}
\fi

\ifARXIV
%\bibliography{master}

\else
\bibliography{master}

\begin{thebibliography}{34}
\providecommand{\natexlab}[1]{#1}
\providecommand{\url}[1]{\texttt{#1}}
\expandafter\ifx\csname urlstyle\endcsname\relax
  \providecommand{\doi}[1]{doi: #1}\else
  \providecommand{\doi}{doi: \begingroup \urlstyle{rm}\Url}\fi

\bibitem[Abid et~al.(2014)Abid, Dal~Zilio, and Le~Botlan]{Abid:2014aa}
N.~Abid, S.~Dal~Zilio, and D.~Le~Botlan.
\newblock A formal framework to specify and verify real-time properties on
  critical systems.
\newblock \emph{International Journal of Critical Computer-Based Systems},
  5\penalty0 (1-2):\penalty0 4--30, 2014.
\newblock \doi{10.1504/IJCCBS.2014.059593}.

\bibitem[Bensalem et~al.(2011)Bensalem, de~Silva, Ingrand, and
  Yan]{Bensalem:2011uf}
S.~Bensalem, L.~de~Silva, F.~Ingrand, and R.~Yan.
\newblock {A Verifiable and Correct-by-Construction Controller for Robot
  Functional Levels}.
\newblock \emph{Journal of Software Engineering for Robotics}, 1\penalty0
  (2):\penalty0 1--19, Sept. 2011.
\newblock URL \url{http://arxiv.org/abs/0908.0221v1}.

\bibitem[Berthomieu and Diaz(1991)]{Berthomieu:1991wv}
B.~Berthomieu and M.~Diaz.
\newblock {Modeling and Verification of Time-Dependent Systems Using Time Petri
  Nets}.
\newblock \emph{IEEE Transactions on Software Engineering}, 17\penalty0
  (3):\penalty0 259--273, Mar. 1991.
\newblock \doi{10.1109/32.75415}.

\bibitem[Berthomieu et~al.(2007)Berthomieu, Bodeveix, Filali, Garavel, Lang,
  Le~Botlan, Vernadat, and Dal~Zilio]{Berthomieu:2007ab}
B.~Berthomieu, J.-P. Bodeveix, M.~Filali, H.~Garavel, F.~Lang, D.~Le~Botlan,
  F.~Vernadat, and S.~Dal~Zilio.
\newblock The syntax and semantics of fiacre.
\newblock Technical Report 7264, LAAS, 2007.
\newblock URL \url{https://projects.laas.fr/fiacre/doc/fiacre.pdf}.

\bibitem[Berthomieu et~al.(2008)Berthomieu, Bodeveix, Farail, Filali, Garavel,
  Gaufillet, Lang, and Vernadat]{Berthomieu:2008vo}
B.~Berthomieu, J.-P. Bodeveix, P.~Farail, M.~Filali, H.~Garavel, P.~Gaufillet,
  F.~Lang, and F.~Vernadat.
\newblock {Fiacre: an Intermediate Language for Model Verification in the
  Topcased Environment}.
\newblock In \emph{Embedded Real-Time Software and Systems}, Toulouse, 2008.
\newblock URL \url{https://hal.laas.fr/inria-00262442}.

\bibitem[Biggar and Zamani(2020)]{Biggar:2020aa}
O.~Biggar and M.~Zamani.
\newblock A framework for formal verification of behavior trees with linear
  temporal logic.
\newblock \emph{IEEE Robotics and Automation Letters}, 5\penalty0 (2):\penalty0
  2341--2348, 2020.
\newblock \doi{10.1109/LRA.2020.2970634}.

\bibitem[Chaki et~al.(2004)Chaki, Clarke, Ouaknine, Sharygina, and
  Sinha]{Chaki:2004aa}
S.~Chaki, E.~M. Clarke, J.~Ouaknine, N.~Sharygina, and N.~Sinha.
\newblock State/event-based software model checking.
\newblock In E.~A. Boiten, J.~Derrick, and G.~Smith, editors, \emph{Integrated
  Formal Methods}, pages 128--147, Berlin, Heidelberg, 2004. Springer Berlin
  Heidelberg.
\newblock ISBN 978-3-540-24756-2.
\newblock \doi{10.1007/978-3-540-2}.

\bibitem[Clarke et~al.(2012)Clarke, Klieber, Nov{\'a}{\v{c}}ek, and
  Zuliani]{Clarke:2012uv}
E.~M. Clarke, W.~Klieber, M.~Nov{\'a}{\v{c}}ek, and P.~Zuliani.
\newblock \emph{Model Checking and the State Explosion Problem}, pages 1--30.
\newblock Springer, Berlin, Heidelberg, 2012.
\newblock ISBN 978-3-642-35746-6.
\newblock \doi{10.1007/978-3-642-35746-6\_1}.

\bibitem[Colledanchise and Natale(2018)]{Colledanchise:2018vt}
M.~Colledanchise and L.~Natale.
\newblock {Improving the Parallel Execution of Behavior Trees}.
\newblock In \emph{IEEE/RSJ International Conference on Intelligent Robots and
  Systems}, Sept. 2018.
\newblock \doi{10.1109/IROS.2018.8593504}.

\bibitem[Colledanchise and Natale(2021)]{Colledanchise:2021aa}
M.~Colledanchise and L.~Natale.
\newblock On the implementation of behavior trees in robotics.
\newblock \emph{IEEE Robotics and Automation Letters}, 6\penalty0 (3):\penalty0
  5929--5936, 2021.
\newblock \doi{10.1109/LRA.2021.3087442}.

\bibitem[Colledanchise and {\"O}gren(2018)]{Colledanchise:2018ub}
M.~Colledanchise and P.~{\"O}gren.
\newblock \emph{Behavior Trees in Robotics and {AI}}.
\newblock {CRC} Press, jul 2018.
\newblock \doi{10.1201/9780429489105}.

\bibitem[Colledanchise et~al.(2017)Colledanchise, Murray, and
  {\"O}gren]{Colledanchise:2017ac}
M.~Colledanchise, R.~M. Murray, and P.~{\"O}gren.
\newblock Synthesis of correct-by-construction behavior trees.
\newblock In \emph{IEEE/RSJ International Conference on Intelligent Robots and
  Systems (IROS)}, pages 6039--6046, 2017.
\newblock \doi{10.1109/IROS.2017.8206502}.

\bibitem[Colledanchise et~al.(2019)Colledanchise, Almeida, and
  {\"O}gren]{Colledanchise:2019vf}
M.~Colledanchise, D.~Almeida, and P.~{\"O}gren.
\newblock Towards blended reactive planning and acting using behavior trees.
\newblock In \emph{International Conference on Robotics and Automation (ICRA)},
  pages 8839--8845. IEEE, 2019.
\newblock \doi{10.1109/ICRA.2019.8794128}.

\bibitem[Colledanchise et~al.(2021)Colledanchise, Cicala, Domenichelli, Natale,
  and Tacchella]{Colledanchise:2021ab}
M.~Colledanchise, G.~Cicala, D.~E. Domenichelli, L.~Natale, and A.~Tacchella.
\newblock Formalizing the execution context of behavior trees for runtime
  verification of deliberative policies.
\newblock In \emph{IEEE/RSJ International Conference on Intelligent Robots and
  Systems (IROS)}, pages 9841--9848, 2021.
\newblock \doi{10.1109/IROS51168.2021.9636129}.

\bibitem[{Dal Zilio} et~al.(2023){Dal Zilio}, Hladik, Ingrand, and
  Mallet]{Dal-Zilio:2023aa}
S.~{Dal Zilio}, P.-E. Hladik, F.~Ingrand, and A.~Mallet.
\newblock A formal toolchain for offline and run-time verification of robotic
  systems.
\newblock \emph{Robotics and Autonomous Systems}, 159:\penalty0 104301, 2023.
\newblock \doi{10.1016/j.robot.2022.104301}.

\bibitem[Ghiorzi and Tacchella(2024)]{Ghiorzi:2024aa}
E.~Ghiorzi and A.~Tacchella.
\newblock Execution semantics of behavior trees in robotic applications.
\newblock \emph{arXiv preprint arXiv:2408.00090}, 2024.
\newblock \doi{10.48550/arXiv.2408.00090}.

\bibitem[Gugliermo et~al.(2024)Gugliermo, {C{\'a}ceres Dom{\'\i}nguez},
  Iannotta, Stoyanov, and Schaffernicht]{Gugliermo:2024aa}
S.~Gugliermo, D.~{C{\'a}ceres Dom{\'\i}nguez}, M.~Iannotta, T.~Stoyanov, and
  E.~Schaffernicht.
\newblock Evaluating behavior trees.
\newblock \emph{Robotics and Autonomous Systems}, 178:\penalty0 104714, 2024.
\newblock \doi{10.1016/j.robot.2024.104714}.

\bibitem[Hladik et~al.(2021)Hladik, Ingrand, Dal~Zilio, and
  Tekin]{Hladik:2021vt}
P.-E. Hladik, F.~Ingrand, S.~Dal~Zilio, and R.~Tekin.
\newblock Hippo: A formal-model execution engine to control and verify critical
  real-time systems.
\newblock \emph{Journal of Systems and Software}, 181:\penalty0 111033, 2021.
\newblock \doi{10.1016/j.jss.2021.111033}.

\bibitem[Ingrand(2024{\natexlab{a}})]{Ingrand:2024aa}
F.~Ingrand.
\newblock Proskill: A formal skill language for acting in robotics.
\newblock Technical report, arXiv, 2024{\natexlab{a}}.
\newblock URL \url{https://arxiv.org/abs/2403.07770}.

\bibitem[Ingrand(2024{\natexlab{b}})]{Ingrand:2024ab}
F.~Ingrand.
\newblock {BT2Fiacre (Behavior Tree 2 Fiacre)}, Oct. 2024{\natexlab{b}}.
\newblock URL \url{https://laas.hal.science/hal-04720141}.

\bibitem[Ingrand and Ghallab(2017)]{Ingrand:2015ue}
F.~Ingrand and M.~Ghallab.
\newblock {Deliberation for autonomous robots: A survey}.
\newblock \emph{Artificial Intelligence}, 247:\penalty0 10--44, June 2017.
\newblock \doi{10.1016/j.artint.2014.11.003}.

\bibitem[Iovino et~al.(2022)Iovino, Scukins, Styrud, {\"O}gren, and
  Smith]{Iovino:2022aa}
M.~Iovino, E.~Scukins, J.~Styrud, P.~{\"O}gren, and C.~Smith.
\newblock A survey of behavior trees in robotics and ai.
\newblock \emph{Robotics and Autonomous Systems}, 154:\penalty0 104096, 2022.
\newblock ISSN 0921-8890.
\newblock \doi{10.1016/j.robot.2022.104096}.

\bibitem[Kl{\"o}ckner(2013)]{Klockner:2013aa}
A.~Kl{\"o}ckner.
\newblock Interfacing behavior trees with the world using description logic.
\newblock In \emph{AIAA Guidance, Navigation, and Control (GNC) Conference},
  2013.
\newblock \doi{10.2514/6.2013-4636}.

\bibitem[K\"{o}ckemann et~al.(2023)K\"{o}ckemann, Calisi, Gemignani, Renoux,
  and Saffiotti]{Kockemann:2023aa}
U.~K\"{o}ckemann, D.~Calisi, G.~Gemignani, J.~Renoux, and A.~Saffiotti.
\newblock Planning for automated testing of implicit constraints in behavior
  trees.
\newblock In \emph{International Conference on Automated Planning and
  Scheduling}, ICAPS '23. AAAI Press, 2023.
\newblock \doi{10.1609/icaps.v33i1.27247}.

\bibitem[Macenski et~al.(2024)Macenski, White, and Wallace]{ros2nav2:2024aa}
S.~Macenski, R.~White, and J.~Wallace.
\newblock Nav2 behavior trees, 2024.
\newblock URL \url{https://docs.nav2.org/behavior_trees/index.html}.

\bibitem[Mart{\'\i}n et~al.(2021)Mart{\'\i}n, Clavero, Matell{\'a}n, and
  Rodr{\'\i}guez]{Martin:2021aa}
F.~Mart{\'\i}n, J.~G. Clavero, V.~Matell{\'a}n, and F.~J. Rodr{\'\i}guez.
\newblock Plansys2: A planning system framework for ros2.
\newblock In \emph{IEEE/RSJ International Conference on Intelligent Robots and
  Systems (IROS)}, pages 9742--9749. IEEE, 2021.
\newblock \doi{10.1109/IROS51168.2021.9636544}.

\bibitem[Marzinotto et~al.(2014)Marzinotto, Colledanchise, Smith, and
  {\"O}gren]{Marzinotto:2014tg}
A.~Marzinotto, M.~Colledanchise, C.~Smith, and P.~{\"O}gren.
\newblock Towards a unified behavior trees framework for robot control.
\newblock In \emph{IEEE International Conference on Robotics and Automation
  (ICRA)}, pages 5420--5427. IEEE, 2014.
\newblock \doi{10.1109/ICRA.2014.6907656}.

\bibitem[Micheli et~al.(2025)Micheli, Bit-Monnot, R{\"o}ger, Scala, Valentini,
  Framba, Rovetta, Trapasso, Bonassi, Gerevini, Iocchi, Ingrand, K{\"o}ckemann,
  Patrizi, Saetti, Serina, and Stock]{Micheli:2025aa}
A.~Micheli, A.~Bit-Monnot, G.~R{\"o}ger, E.~Scala, A.~Valentini, L.~Framba,
  A.~Rovetta, A.~Trapasso, L.~Bonassi, A.~E. Gerevini, L.~Iocchi, F.~Ingrand,
  U.~K{\"o}ckemann, F.~Patrizi, A.~Saetti, I.~Serina, and S.~Stock.
\newblock Unified planning: Modeling, manipulating and solving ai planning
  problems in python.
\newblock \emph{SoftwareX}, 29:\penalty0 102012, 2025.
\newblock \doi{10.1016/j.softx.2024.102012}.

\bibitem[{\"O}gren and Sprague(2022)]{Ogren:2022aa}
P.~{\"O}gren and C.~I. Sprague.
\newblock Behavior trees in robot control systems.
\newblock \emph{Annual Review of Control, Robotics, and Autonomous Systems},
  5\penalty0 (Volume 5, 2022):\penalty0 81--107, 2022.
\newblock \doi{10.1146/annurev-control-042920-095314}.

\bibitem[Quigley et~al.(2009)Quigley, Gerkey, Conley, Faust, Foote, Leibs,
  Berger, Wheeler, and Ng]{Quigley:2009tg}
M.~Quigley, B.~Gerkey, K.~Conley, J.~Faust, T.~Foote, J.~Leibs, E.~Berger,
  R.~Wheeler, and A.~Ng.
\newblock {ROS: an open-source Robot Operating System}.
\newblock In \emph{ICRA Workshop on Open Source Software}. Kobe, Japan, 2009.
\newblock URL \url{http://robotics.stanford.edu/~ang/papers/icraoss09-ROS.pdf}.

\bibitem[Schulz-Rosengarten et~al.(2024)Schulz-Rosengarten, Ahmad, Clement, von
  Hanxleden, Asch, Lohstroh, Lee, Quiros, and
  Shukla]{Schulz-Rosengarten:2024aa}
A.~Schulz-Rosengarten, A.~Ahmad, M.~Clement, R.~von Hanxleden, B.~Asch,
  M.~Lohstroh, E.~A. Lee, G.~Quiros, and A.~Shukla.
\newblock Behavior trees with dataflow: Coordinating reactive tasks in lingua
  franca.
\newblock In \emph{Proceedings of the 2024 IEEE/ACM 46th International
  Conference on Software Engineering: Companion Proceedings}, ICSE-Companion
  '24, pages 304--305, New York, NY, USA, 2024. Association for Computing
  Machinery.
\newblock \doi{10.1145/3639478.3643093}.

\bibitem[Serbinowska et~al.(2024)Serbinowska, Potteiger, Tumlin, and
  Johnson]{Serbinowska:2024ab}
S.~S. Serbinowska, N.~Potteiger, A.~M. Tumlin, and T.~T. Johnson.
\newblock Verification of behavior trees with contingency monitors.
\newblock \emph{Electronic Proceedings in Theoretical Computer Science},
  411:\penalty0 56--72, Nov. 2024.
\newblock \doi{10.4204/eptcs.411.4}.

\bibitem[Street et~al.(2024)Street, Warsame, Mansouri, Klauck, Henkel,
  Lampacrescia, Palmas, Lange, Ghiorzi, Tacchella, Azrou, Lallement, Morelli,
  {I. Chen}, Wallis, Bernagozzi, Rosa, Randazzo, Faraci, and
  Natale]{Street:2024aa}
C.~Street, Y.~Warsame, M.~Mansouri, M.~Klauck, C.~Henkel, M.~Lampacrescia,
  M.~Palmas, R.~Lange, E.~Ghiorzi, A.~Tacchella, R.~Azrou, R.~Lallement,
  M.~Morelli, G.~{I. Chen}, D.~Wallis, S.~Bernagozzi, S.~Rosa, M.~Randazzo,
  S.~Faraci, and L.~Natale.
\newblock Towards a verifiable toolchain for robotics.
\newblock In \emph{AAAI Fall Symposium Series}, AAAI Symposium Series (Fall).
  AAAI Press, Aug. 2024.
\newblock \doi{10.1609/aaaiss.v4i1.31823}.

\bibitem[Wang et~al.(2024)Wang, Dai, Zhao, Zhang, and Bliudze]{Wang:2024aa}
Q.~Wang, H.~Dai, Y.~Zhao, M.~Zhang, and S.~Bliudze.
\newblock Enabling behaviour tree verification via a translation to bip.
\newblock In D.~Marmsoler and M.~Sun, editors, \emph{Formal Aspects of
  Component Software}, pages 3--20, Cham, 2024. Springer Nature Switzerland.
\newblock \doi{10.1007/978-3-031-71261-6\_1}.

\end{thebibliography}
\fi

\newpage
\appendix
\section{A BT \btn{Action} \hfiacre{} process}

\label{app:app1}
\begin{lstlisting}[caption={The \hfiacre{} process specification of the \btn{takeoff} BT \btn{Action} node.}, numbers=left, xleftmargin=15pt, label={lst:factionh}, language=fiacre]
process btnode_takeoff_btn21 (&btnode: btnode_array, &fls: sv_fls, &battery: sv_battery) is

states start_, tick_node, success, failure, halt, halted, running, error,
   Action_takeoff, dispatch, Action_takeoff_sync, done

var callb: bool,
   report_halted:bool := false, ret_val: ret_status

from start_
   wait [0,0];
   on (btnode[takeoff_btn21].caller <> None); // Wait until we are called
   report_halted := false;
   if (btnode[takeoff_btn21].rstatus = halt_me) then // are we being instructed to halt?
       report_halted := true;
       to halt
   end;
   // Btnode Action 'takeoff_btn21' has been called (:height 1.000000  :duration 0 ) 
   btnode[takeoff_btn21].rstatus := no_ret_status; // just initializing the rstatus
to tick_node

from halt
   wait [0,0];
   // Btnode Action 'takeoff_btn21' is being halted (:height 1.000000  :duration 0 ) 
   callb := Fiacre_Action_takeoff_halt (btnode[takeoff_btn21]); //This call the external which
to halted // halts the action

from tick_node
   wait [0,0];
   to Action_takeoff

from Action_takeoff
   // synthesized action arg index  2
   btnode[takeoff_btn21].ArgIndex := 2;
   // Btnode Action 'takeoff_btn21' calling its action (start task)
   start Fiacre_Action_takeoff_task (btnode[takeoff_btn21]); // this call the Fiacre task
to Action_takeoff_sync // which handles this action.

from Action_takeoff_sync
   sync Fiacre_Action_takeoff_task ret_val; // wait until the Fiacre task returns
   // Btnode Action 'takeoff_btn21' returned (sync task)
   to dispatch

from dispatch
   wait [0,0]; // we dispatch to the proper state according to the return values
   if (ret_val = success) then
       to success
   elsif (ret_val = failure) then
       to failure
   elsif (ret_val = running) then
       to running
   else
       to error // a priori unreachable
   end

from success
   wait [0,0];
   // Btnode Action 'takeoff_btn21' (:height 1.000000  :duration 0 ) returns success.
   btnode[takeoff_btn21].rstatus := success;
   to done

from failure
   wait [0,0];
   if (report_halted) then // this is mostly for traces
       // Btnode Action 'takeoff_btn21' (:height 1.000000  :duration 0 ) returns halted failure.
       null
   else
       // Btnode Action 'takeoff_btn21' (:height 1.000000  :duration 0 ) returns failure.
       null
   end;
   btnode[takeoff_btn21].rstatus := failure;
   to done

from halted
   wait [0,0];
   // Btnode Action 'takeoff_btn21' (:height 1.000000  :duration 0 ) has been halted.
   report_halted := true;
   to failure

from running
   wait [0,0];
   // Btnode Action 'takeoff_btn21' (:height 1.000000  :duration 0 ) returns running.
   btnode[takeoff_btn21].rstatus := running;
   to done

from done
   wait [0,0];
   // Btnode Action 'takeoff_btn21' returning control to caller and back to 'start_'
   btnode[takeoff_btn21].caller := None; // we relinquish the tick and go back waiting
to start_
\end{lstlisting}

\section{A BT \btn{Sequence}  \hfiacre{} process}

\label{app:app2}
\begin{lstlisting}[caption={The \hfiacre{} process specification of a BT \btn{Sequence} node.}, numbers=left, xleftmargin=15pt, label={lst:fsequenceh}, language=fiacre]
process btnode_Sequence13_btn6 (&btnode: btnode_array) is

states start_, tick_node, success, failure, halt, halted, halt_wait, running, error,
    Fallback15_btn7, Fallback15_btn7_done, Eval25_btn12, Eval25_btn12_done,
    Fallback27_btn13, Fallback27_btn13_done, done

var next_seq: 1..3 := 1

from start_
    wait [0,0];
    on (btnode[Sequence13_btn6].caller <> None);
    if (btnode[Sequence13_btn6].rstatus = halt_me) then to halt end;
    // Btnode Sequence 'Sequence13_btn6' has been called 
    btnode[Sequence13_btn6].rstatus := no_ret_status;
to tick_node

from halt
    wait [0,0];
    // Btnode Sequence 'Sequence13_btn6' is being halted 
    if (btnode[Fallback15_btn7].rstatus = running) then
        // Btnode Sequence 'Sequence13_btn6' halting Fallback 'Fallback15_btn7'
        btnode[Fallback15_btn7].rstatus := halt_me;
        btnode[Fallback15_btn7].caller := caller_Sequence13_btn6;
        to halt_wait
    end;
    if (btnode[Eval25_btn12].rstatus = running) then
        // Btnode Sequence 'Sequence13_btn6' halting Eval 'Eval25_btn12'
        btnode[Eval25_btn12].rstatus := halt_me;
        btnode[Eval25_btn12].caller := caller_Sequence13_btn6;
        to halt_wait
    end;
    if (btnode[Fallback27_btn13].rstatus = running) then
        // Btnode Sequence 'Sequence13_btn6' halting Fallback 'Fallback27_btn13'
        btnode[Fallback27_btn13].rstatus := halt_me;
        btnode[Fallback27_btn13].caller := caller_Sequence13_btn6;
        to halt_wait
    end;
    to halted

from halt_wait
    on ((btnode[Fallback15_btn7].caller = None) and
        (btnode[Eval25_btn12].caller = None) and
        (btnode[Fallback27_btn13].caller = None));
    to halted

from tick_node
    wait [0,0];
    if (next_seq = 1) then to Fallback15_btn7 end;
    if (next_seq = 2) then to Eval25_btn12 end;
    if (next_seq = 3) then to Fallback27_btn13 end;
    to error

from Fallback15_btn7
    wait [0,0];
    // Btnode Sequence 'Sequence13_btn6' calling Fallback 'Fallback15_btn7'
    btnode[Fallback15_btn7].caller := caller_Sequence13_btn6;
    to Fallback15_btn7_done

from Fallback15_btn7_done
    wait [0,0];
    on (btnode[Fallback15_btn7].caller = None);
    // Btnode Sequence 'Sequence13_btn6' getting control back from Fallback 'Fallback15_btn7'
    if (btnode[Fallback15_btn7].rstatus = success) then
        to Eval25_btn12
    elsif (btnode[Fallback15_btn7].rstatus = failure) then
        next_seq := 1;
        to failure
    elsif (btnode[Fallback15_btn7].rstatus = running) then
        next_seq := 1;
        to running
    else
        to error
    end

from Eval25_btn12
    wait [0,0];
    // Btnode Sequence 'Sequence13_btn6' calling Eval 'Eval25_btn12'
    btnode[Eval25_btn12].caller := caller_Sequence13_btn6;
    to Eval25_btn12_done

from Eval25_btn12_done
    wait [0,0];
    on (btnode[Eval25_btn12].caller = None);
    // Btnode Sequence 'Sequence13_btn6' getting control back from Eval 'Eval25_btn12'
    if (btnode[Eval25_btn12].rstatus = success) then
        to Fallback27_btn13
    elsif (btnode[Eval25_btn12].rstatus = failure) then
        next_seq := 1;
        to failure
    elsif (btnode[Eval25_btn12].rstatus = running) then
        next_seq := 2;
        to running
    else
        to error
    end

from Fallback27_btn13
    wait [0,0];
    // Btnode Sequence 'Sequence13_btn6' calling Fallback 'Fallback27_btn13'
    btnode[Fallback27_btn13].caller := caller_Sequence13_btn6;
    to Fallback27_btn13_done

from Fallback27_btn13_done
    wait [0,0];
    on (btnode[Fallback27_btn13].caller = None);
    // Btnode Sequence 'Sequence13_btn6' getting control back from Fallback 'Fallback27_btn13'
    if (btnode[Fallback27_btn13].rstatus = success) then
        next_seq := 1;
        to success
    elsif (btnode[Fallback27_btn13].rstatus = failure) then
        next_seq := 1;
        to failure
    elsif (btnode[Fallback27_btn13].rstatus = running) then
        next_seq := 3;
        to running
    else
        to error
    end

from success
    wait [0,0];
    // Btnode Sequence 'Sequence13_btn6' returns success.
    btnode[Sequence13_btn6].rstatus := success;
    to done

from failure
    wait [0,0];
    // Btnode Sequence 'Sequence13_btn6' returns failure.
    btnode[Sequence13_btn6].rstatus := failure;
    to done

from halted
    wait [0,0];
    // Btnode Sequence 'Sequence13_btn6' has been halted.
    btnode[Sequence13_btn6].rstatus := failure;
    to done

from running
    wait [0,0];
    // Btnode Sequence 'Sequence13_btn6' returns running.
    btnode[Sequence13_btn6].rstatus := running;
    to done

from done
    wait [0,0];
    // Btnode Sequence 'Sequence13_btn6' returning control to caller and back to '_start'
    btnode[Sequence13_btn6].caller := None;
to start_
\end{lstlisting}

\clearpage
\section{\btn{Fallback} and \btn{Parallel} nodes transformation in \fiacre{}}

\begin{figure*}[!ht]
\begin{center}
\includegraphics[angle=-90,origin=c,height=0.6\textheight]{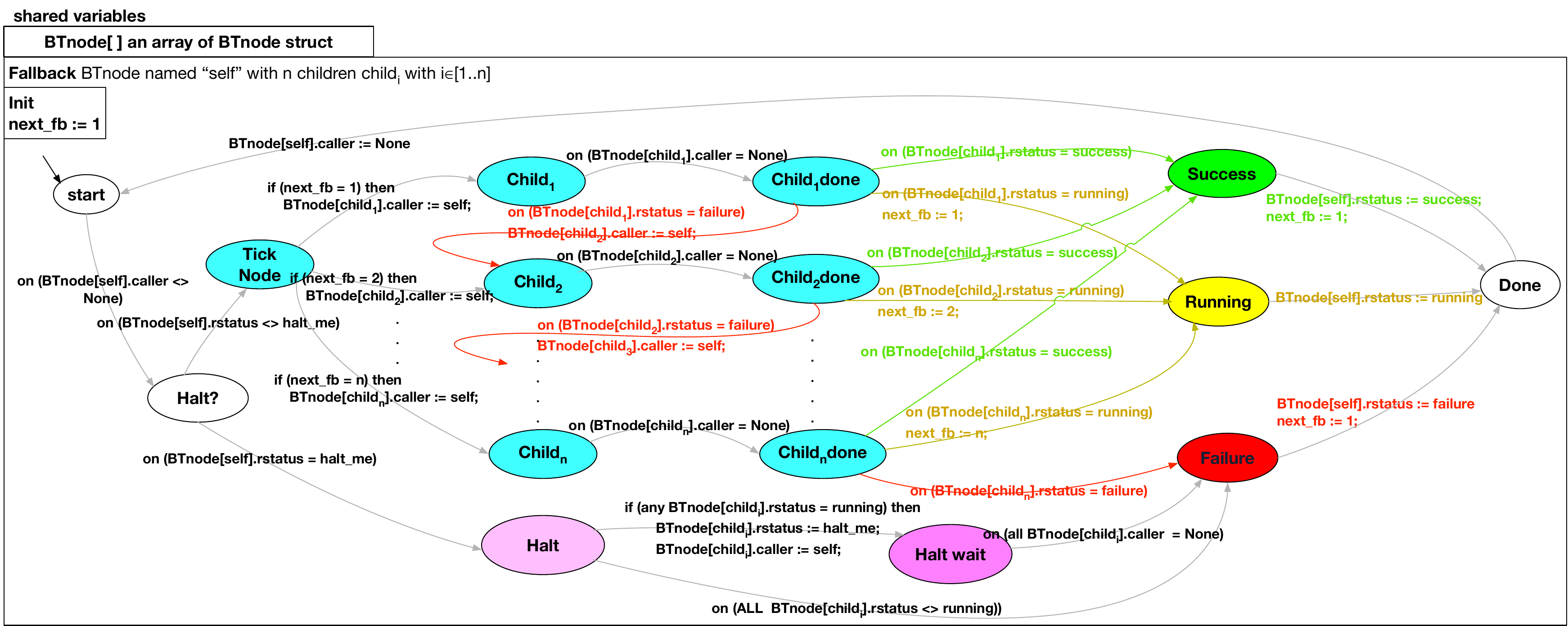}
\caption{The \fiacre{} process modeling the \btn{Fallback} node.}
\label{fig:fallback}
\end{center}
\end{figure*}
  
\begin{figure*}[!ht]
\begin{center}
\includegraphics[angle=-90,origin=c,height=0.6\textheight]{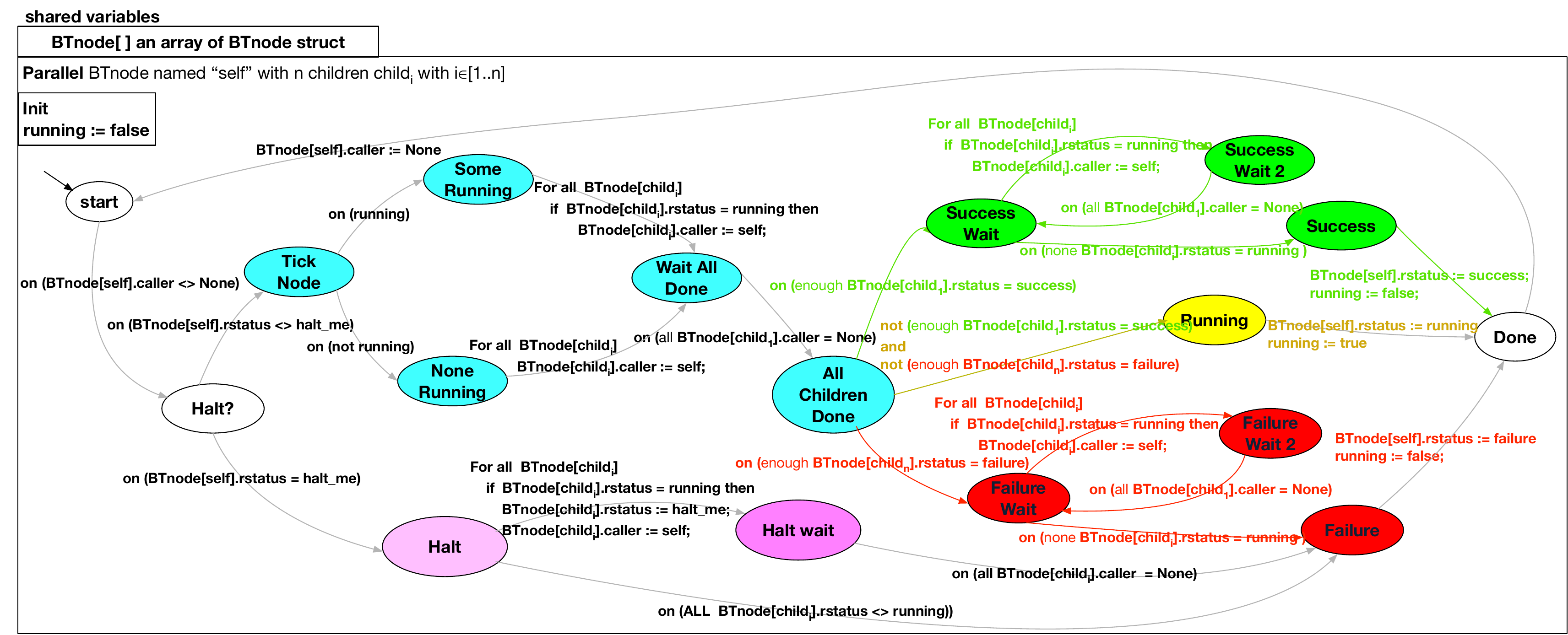}
\caption{The \fiacre{} process modeling the \btn{Parallel} node.}
\label{fig:parallel}
\end{center}
\end{figure*}

\clearpage
\section{The Drone BT.}

\begin{figure*}[!ht]
\begin{center}
\includegraphics[width=0.97\textwidth]{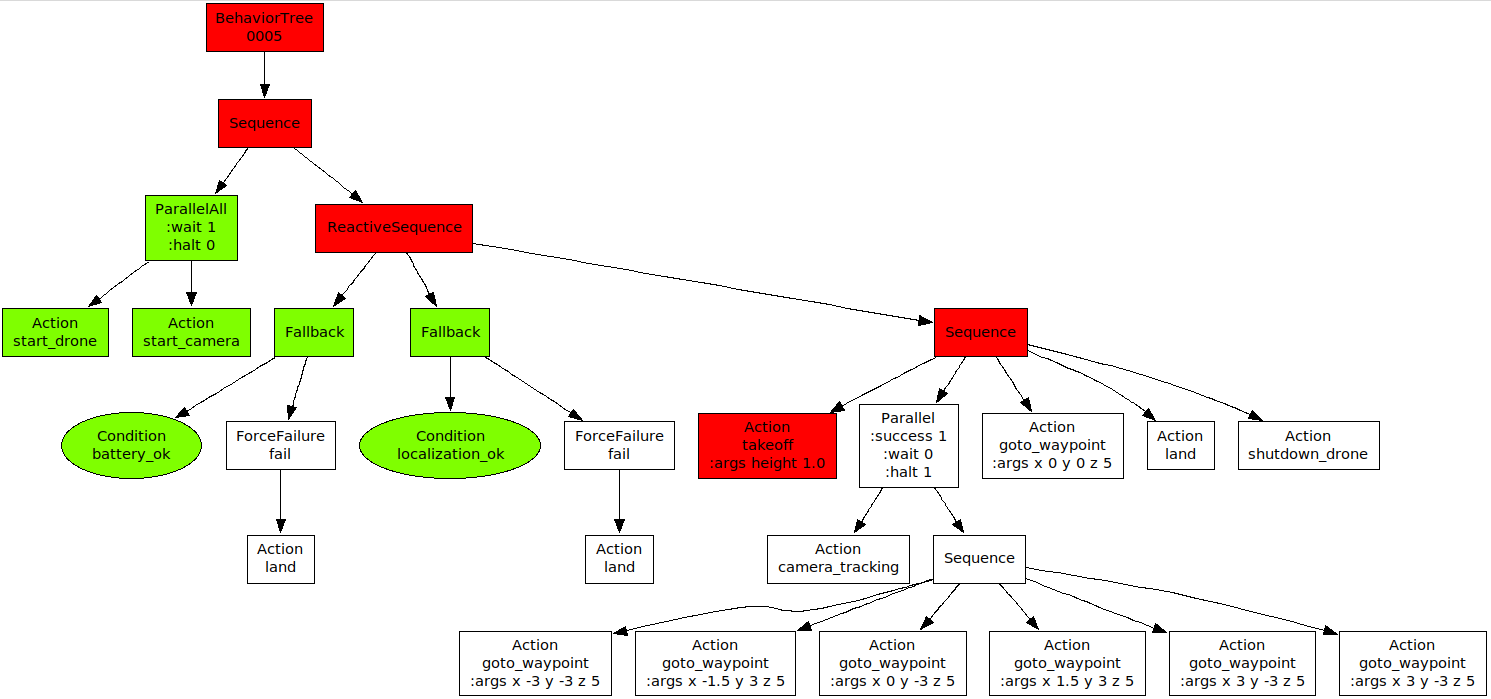}
\caption{The graphical representation of the BT Listing~\ref{lst:dronebt}\code{p}\pageref{lst:dronebt}.}
\label{fig:screen-bt}
\end{center}
\end{figure*}

%\clearpage
\section{A ROS2 Nav2 BT and its BTF equivalent.}

\label{app:ros2nav2}
\begin{lstlisting}[caption={One of the ROS2 Nav2 Behavior Tree (Navigate To Pose With Replanning and Recovery)~\cite{ros2nav2:2024aa}.}, numbers=left, xleftmargin=15pt, label={lst:ros2nav2xml}, language=XML]
<root main_tree_to_execute="MainTree">
  <BehaviorTree ID="MainTree">
    <RecoveryNode number_of_retries="6" name="NavigateRecovery">
      <PipelineSequence name="NavigateWithReplanning">
        <RateController hz="1.0">
          <RecoveryNode number_of_retries="1" name="ComputePathToPose">
            <ComputePathToPose goal="{goal}" path="{path}" planner_id="GridBased"/>
            <ReactiveFallback name="ComputePathToPoseRecoveryFallback">
              <GoalUpdated/>
              <ClearEntireCostmap name="ClearGlobalCostmap-Context" 
                                service_name="global_costmap/clear_entirely_global_costmap"/>
            </ReactiveFallback>
          </RecoveryNode>
        </RateController><
        <RecoveryNode number_of_retries="1" name="FollowPath">
          <FollowPath path="{path}" controller_id="FollowPath"/>
          <ReactiveFallback name="FollowPathRecoveryFallback">
            <GoalUpdated/>
            <ClearEntireCostmap name="ClearLocalCostmap-Context" 
                                service_name="local_costmap/clear_entirely_local_costmap"/>
          </ReactiveFallback>
        </RecoveryNode>
      </PipelineSequence>
      <ReactiveFallback name="RecoveryFallback">
        <GoalUpdated/>
        <RoundRobin name="RecoveryActions">
          <Sequence name="ClearingActions">
            <ClearEntireCostmap name="ClearLocalCostmap-Subtree" 
                                service_name="local_costmap/clear_entirely_local_costmap"/>
            <ClearEntireCostmap name="ClearGlobalCostmap-Subtree" 
                                service_name="global_costmap/clear_entirely_global_costmap"/>
          </Sequence>
          <Spin spin_dist="1.57"/>
          <Wait wait_duration="5"/>
          <BackUp backup_dist="0.15" backup_speed="0.025"/>
        </RoundRobin>
      </ReactiveFallback>
    </RecoveryNode>
  </BehaviorTree>
</root>
\end{lstlisting}

\begin{lstlisting}[caption={The \code{.btf} version of the ROS2 Nav2 BT above (Listing~\ref{lst:ros2nav2xml}\code{p}\pageref{lst:ros2nav2xml}).}, numbers=left, xleftmargin=15pt, label={lst:ros2nav2btf}, language={[btf]Lisp}]
((BehaviorTree :ID MainTree ; The top level root node
     (Recovery :num_retries 6 :name NavigateRecovery
         (PipelineSequence :name NavigateWithReplanning
          (RateController :args (hz 1)
              (Recovery :num_retries 1 :name ComputePathToPose
               (Action :ID ComputePathToPose :args (goal $goal path $path planner_id GridBased))
               (ReactiveFallback :name ComputePathToPoseRecoveryFallback
                (Condition :ID GoalUpdated)
                (Action :ID ClearEntireCostmap :name ClearGlobalCostmap_Context1
                        :args (service_name global_costmap/clear_entirely_global_costmap)))))
          (Recovery :num_retries 1 :name FollowPath
              (Action :ID FollowPath :args (path $path controller_id FollowPath))
              (ReactiveFallback :name FollowPathRecoveryFallback
               (Condition :ID GoalUpdated)
               (Action :ID ClearEntireCostmap :name ClearLocalCostmap_Context2
                       :args (service_name local_costmap/clear_entirely_local_costmap)))))
         (ReactiveFallback :name RecoveryFallback
          (Condition :ID GoalUpdated)
          (RoundRobin :name RecoveryActions
              (Sequence :name ClearingActions
               (Action :ID ClearEntireCostmap :name ClearLocalCostmap_Subtree3 
                       :args ( service_name local_costmap/clear_entirely_local_costmap))
               (Action :ID ClearEntireCostmap :name ClearGlobalCostmap_Subtree4 
                       :args (service_name global_costmap/clear_entirely_global_costmap)))
              (Action :ID Spin :args (spin_dist 1.57))
              (Action :ID Wait :args (wait_duration 5))
              (Action :ID BackUp :args (backup_dist 0.15 backup_speed 0.025)))))))
\end{lstlisting}

\begin{figure*}[!ht]
\begin{center}
\centerline{\includegraphics[width=1.1\textwidth]{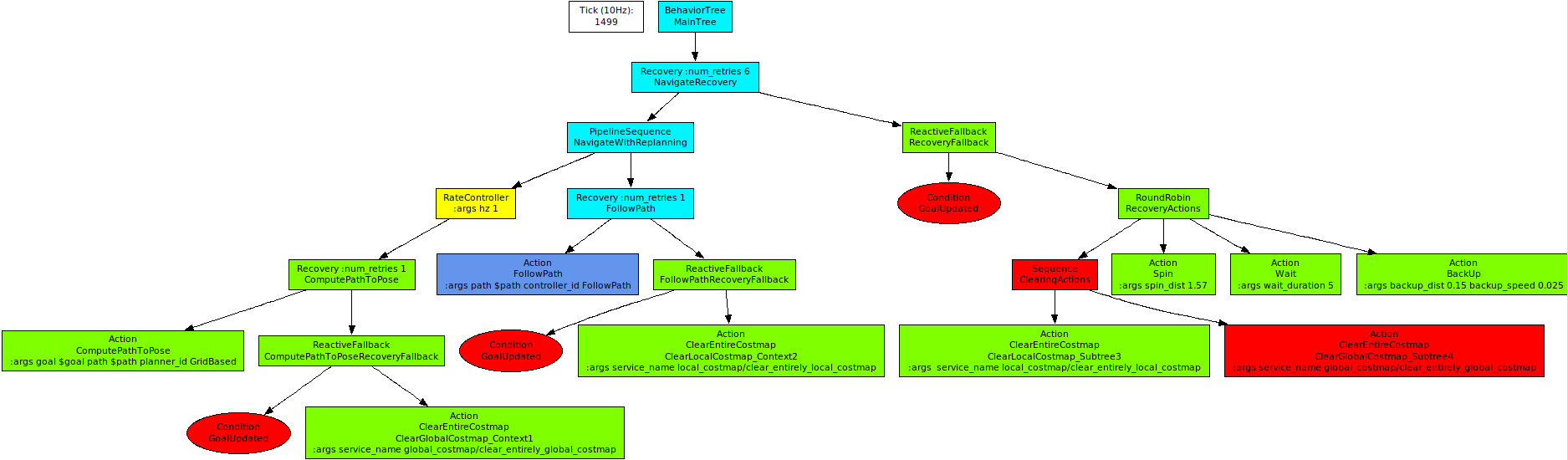}}
\caption{The screen dump of the Nav2 BT (See Listing~\ref{lst:ros2nav2btf}\code{p}\pageref{lst:ros2nav2btf} executing.}
\label{fig:screen-nav2}
\end{center}
\end{figure*}
 
\clearpage
\section{The Mars rover example from~\cite{Biggar:2020aa,Wang:2024aa}.}
\label{app:mars}

\begin{lstlisting}[caption={The BTF model of the Mars rover example from~\cite{Biggar:2020aa,Wang:2024aa}.}, numbers=left, xleftmargin=15pt, label={lst:mars}, language={[btf]Lisp}]
( ;example from 
  ; A framework for formal verification of behavior trees with linear temporal logic.(2020)
  ; O. Biggar and M. Zamani.
  ; also presented in 
  ; Enabling Behaviour Tree Verification via a Translation to BIP. (2024)
  ; Q Wang, H Dai, Y Zhao, M Zhang, S Bliudze
  (defsv meteo ; this defines the meteo state variable
    :states (MInit Normal Storm) ; MInit an undefined init state
    :init MInit  ; the following transitions forbid coming back to MInit
    :transitions ((MInit Normal) (MInit Storm) (Storm Normal) (Normal Storm)))

  (defsv battery
    :states (BInit Good Low) ; BInit just to says that we do not know
    :init BInit  ; the following transitions forbid coming back to BInit
    :transitions ((BInit Good) (BInit Low) (Low Good) (Good Low)))

  (defsv panel
    :states (PInit Folded Unfolded) ; PInit just to says that we do not know
    :init PInit  ; the following transitions forbid coming back to PInit
    :transitions ((PInit Folded) (PInit Unfolded) (Unfolded Folded) (Folded Unfolded)))

(BehaviorTree :name mars_rover
  (Fallback
   (Sequence
    (Eval (= battery Low)) ; if the battery is low
    (Action :ID UnfoldPanels :name unfold_panels :SF) ; we unfold the panel
    (Eval (:= panel Unfolded))) ; and the panel SV becomes unfolded
   (Sequence
    (Eval (= meteo Storm)) ; if the meteo is a storm
    (Action :ID Hibernate :name hibernate :SF) ; we hibernate
    (Eval (:= panel Folded))) ; and we fold the panel
   (Sequence
    (Action :ID DataReady :name dataready :SF) 
    (Action :ID Send :name send :SF)))))

; prove absent ( mars_rover/3/state Unfolded and mars_rover/1/state Storm ) 
\end{lstlisting}

\end{document}